\documentclass[twoside]{article}
\usepackage{amsfonts}
\usepackage{graphicx}
\usepackage[preprint]{aistats2026}
\usepackage[round]{natbib}
\usepackage{fontawesome5}

\usepackage[colorlinks = true,
            linkcolor = magenta,
            urlcolor  = magenta,
            citecolor = magenta,
            anchorcolor = magenta]{hyperref}
\usepackage{url}            % simple URL typesetting
\usepackage{booktabs}       % professional-quality tables
\usepackage{amsfonts}       % blackboard math symbols
\usepackage{nicefrac}       % compact symbols for 1/2, etc.
\usepackage{microtype}      % microtypography       
\usepackage[table]{xcolor}  % colors
\usepackage{graphicx}
\usepackage{subcaption}
\usepackage{amsmath}
\usepackage{amssymb}
\usepackage{mathtools}
\usepackage{amsthm}
\usepackage{makecell}
\theoremstyle{plain}
\newtheorem{theorem}{Theorem}[section]

\newtheorem{lemma}[theorem]{Lemma}
\newtheorem{corollary}[theorem]{Corollary}
\theoremstyle{definition}

\theoremstyle{remark}

\usepackage[textsize=tiny]{todonotes}
\usepackage{tikz-qtree}
\usepackage{tikz}
\usetikzlibrary{positioning, arrows.meta}
\usepackage{amsfonts}
\usepackage{lipsum}
\usepackage{mathtools}
\usepackage{multirow}
\usepackage{algorithm}
\usepackage{algpseudocode}
\usepackage{enumitem}
\usepackage{dsfont}

\newcommand{\indicator}[1]{\mathds{1}_{\{#1\}}}

\newcommand{\M}{\mathcal{M}}
\newcommand{\T}{\mathcal{T}}
\newcommand{\B}{\mathcal{B}}
\newcommand{\C}{\mathcal{C}}
\renewcommand{\S}{\mathcal{S}}
\newcommand{\A}{\mathcal{A}}
\newcommand{\st}{\tilde{s}}
\newcommand{\s}{sense}
\newcommand{\bl}{blind}
\newcommand{\Vn}{V_{\M_{k,N}}}
\newcommand{\V}{V_{\M_k}}

\newcommand{\ignore}[1]{}
\allowdisplaybreaks[4]

\newcommand{\appendixcontents}{
    \begin{center}
        \textbf{Appendix Contents}
    \end{center}
    \noindent
    \hyperref[sec:proofs]{\textcolor{blue}{A. Proofs}} \dotfill \pageref{sec:proofs} \\
   \hspace*{1em} \hyperref[proof:sense_thr]{\textcolor{blue}{A.1 Proof of Theorem~\ref{thm:always_sense}}} \dotfill \pageref{proof:sense_thr} \\
   \hspace*{1em} \hyperref[proof:depth]{\textcolor{blue}{A.2 Proof of Theorem~\ref{thm:depththreshold}}} \dotfill \pageref{proof:depth} \\
 \hspace*{1em} \hyperref[proof:onestep]{\textcolor{blue}{A.3 Proof of Lemma~\ref{thm:one-step}}} \dotfill \pageref{proof:onestep} \\
   \hspace*{1em} \hyperref[proof:pi_N_optimal]{\textcolor{blue}{A.4 Proof of Theorem~\ref{thm:pi_N_optimal}}} \dotfill \pageref{proof:pi_N_optimal} \\
   \hspace*{1em} \hyperref[proof:epsilon]{\textcolor{blue}{A.5 Proof of Lemma~\ref{thm:epsilon}}} \dotfill \pageref{proof:epsilon} \\[0.25em]
\hyperref[sec:Experimental Design]{\textcolor{blue}{B. Experimental Design and Setup}} \dotfill \pageref{sec:Experimental Design} \\
    \hyperref[sec:SPI Analysis]{\textcolor{blue}{C. Analysis of SPI}} \dotfill \pageref{sec:SPI Analysis} \\
    \hyperref[sec:ATM Analysis]{\textcolor{blue}{D. Analysis of ATM Heuristic}} \dotfill \pageref{sec:ATM Analysis} \\
    \hyperref[sec:addl-results]{\textcolor{blue}{E. Additional Results}} \dotfill \pageref{sec:addl-results} \\
    \hyperref[sec:non-uniform]{\textcolor{blue}{F. Extension to Non-Uniform Sensing Costs}} \dotfill \pageref{sec:non-uniform} \\
    \hyperref[Inventory]{\textcolor{blue}{G. Inventory Management Case Study}} \dotfill \pageref{Inventory}
}

\definecolor{myblue}{RGB}{224, 240, 255}
\begin{document}



\twocolumn[

\aistatstitle{MDPs with a State Sensing Cost}

\aistatsauthor{ Vansh Kapoor \And Jayakrishnan Nair}

\aistatsaddress{IIT Bombay\\vanshk@cs.cmu.edu \And IIT Bombay\\jayakrishnan.nair@ee.iitb.ac.in} ]

\begingroup
\renewcommand{\thefootnote}{}

\footnotetext{
\faGithub\ Code and implementation details are available at \\
\texttt{\href{https://github.com/Vansh28Kapoor/POMDPs-With-Sensing-Cost}{https://github.com/Vansh28Kapoor/POMDPs-With-Sensing-Cost}}
}

\endgroup
% \begingroup
% \renewcommand{\thefootnote}{}
% \footnotetext{Code and implementation details are available \href{https://github.com/Vansh28Kapoor/POMDPs-With-Sensing-Cost}{here}.}
% \endgroup
\begin{abstract}
  In many practical sequential decision-making problems, tracking the state of the environment incurs a sensing/computation cost. In these settings, the agent's interaction with its environment includes the additional component of deciding \emph{when} to sense the state, in a manner that balances the value associated with optimal (state-specific) actions and the cost of sensing. We formulate this as an expected discounted cost Markov Decision Process (MDP), wherein the agent incurs an additional cost for sensing its next state, but has the option to take actions while remaining `blind' to the system state.
    We pose this problem as a classical discounted cost MDP with an expanded (countably infinite) state space.
    While computing the optimal policy for this MDP is intractable in general, we derive lower bounds on the optimal value function, which allow us to bound the suboptimality gap of any policy. 
    We also propose a computationally efficient algorithm SPI, based on policy improvement, which in practice performs close to the optimal policy. Finally, we benchmark against the state-of-the-art via a numerical case study.
\end{abstract}
\vspace{-0.2cm}
\section{Introduction}
\vspace{-0.2cm}
Markov Decision Processes (MDPs) provide a powerful framework for modeling sequential interactions between an agent and an adaptive environment that `responds' to the agent's actions. In the classical MDP formulation, at each time step $t$, the agent sees the environment's state, denoted $X_t$, and selects an action $A_t$. This action generates a feedback signal (cost) determined by the state-action pair $ (X_t, A_t)$, and triggers a random, action-dependent Markovian transition in the state. However, in many applications, `seeing' the current state involves a cost. For example:

$\bullet$ \textbf{Healthcare:} Monitoring a patient’s state during an ongoing intervention, such as white blood cell counts in anti-HIV therapy or lab tests for ICU patients, incurs monetary or delay costs \citep{HIV}.\\
$\bullet$ \textbf{Mobile Applications:} Sensing a user’s location, motion, or environment on mobile devices incurs energy costs, which must be balanced with user experience goals \citep{mobilemarkov}. \\
$\bullet$ \textbf{Wireless Sensor Networks:} Sensing the network state might require turning on battery-operated (and therefore energy-constrained) sensors.\\
$\bullet$ \textbf{Distributed Sensing:} Aggregating sensor data to determine system state involves computational costs.
$\bullet$ \textbf{Remote Surveillance:} Transmitting state information (in the form of images/video) to a controller incurs communication costs.\\
$\bullet$ \textbf{Robotics:} In applications where an autonomous robot is engaged in a task, sensing the surroundings might induce a latency (cost) in task completion.

This motivates integrating the cost of state sensing and a model for opportunistic state sensing into the MDP framework. However, such integration introduces technical challenges. In particular, if the agent chooses not to sense the system state (while continuing to interact with the environment), it is left with a \emph{belief distribution} over the system state. Incorporating this belief into the MDP formulation induces a state space explosion, which makes the decision problem intractable.

In this paper, we formulate a cost-sensing MDP built on top of a finite, infinite-horizon discounted cost baseline MDP. The `augmented' MDP, which incorporates the state sensing cost, has a countably infinite state space. At each time/epoch, the agent must, in addition to taking an action, decide whether or not to sense the \emph{next} state of the MDP. If it decides to sense, it incurs an additional state sensing cost in that epoch. There is thus a non-trivial trade-off between the cost induced by suboptimal actions (under state uncertainty) and the cost of state sensing.

While the `augmented' MDP admits a stationary Markov policy, it is computationally intractable given the infinite state space. The goal of this paper is to design tractable, near-optimal algorithms for solving it. Our key contributions are as follows:

$\bullet$ We propose SPI, a policy-improvement-based algorithm that selectively searches for improving blind action sequences and is near-optimal in practice.\\
$\bullet$ We analyse a sequence of truncated (and therefore finite) MDPs that restrict the number of consecutive non-state-sensing (a.k.a., blind) actions the agent can take. We provide a sufficient condition for the optimal policy under such a truncated MDP to also be optimal for the original (infinite state space) MDP of interest.\\
$\bullet$ We utilize the truncated MDPs to derive lower bounds on the optimal value function, enabling computation of the suboptimality gap for any policy.\\
$\bullet$ We derive a computable state-sensing cost threshold, below which always sensing is optimal.\\
$\bullet$ Finally, we conduct an extensive numerical case study,  including a real-world healthcare application, demonstrating that SPI consistently outperforms state-of-the-art POMDP solvers under diverse sensing costs, while maintaining reasonable compute time.

It is worth noting that our formulation can be posed as a Partially Observable Markov Decision Processes (POMDP) and indeed we benchmark our methods against state-of-the-art POMDP solvers (see Section~\ref{sec:numerics}). However, in doing so, one loses the specific problem structure that arises in the opportunistic state sensing formulation, which we seek to exploit here for computational tractability. Since POMDPs are known to be intractable in general \citep{POMDPcomplexity}, leveraging this structure is crucial. Specifically, our algorithms and analytical results exploit the specific structure of this formulation, namely that the uncertainty in the belief distribution can be collapsed entirely by paying a sensing cost, allowing a good policy to go `blind' until the cost induced by state uncertainty is outweighed by the sensing cost. A generic MDP/POMDP solver would not be able to exploit this structure. Moreover, our algorithmic design and proof arguments rely on classical MDP tools--policy improvement and the Bellman operator--whose POMDP analogues are significantly weaker.

\textbf{Related Literature.} Formulations equivalent to ours have been analysed in \citep{hansen1994cost,bellinger2021active,nam2021reinforcement,ATM}; the last two references refer to this formulation as an \emph{Action-Contingent-Noiseless-Observable MDP}, or ACNO-MDP. \citet{hansen1994cost} proposes a truncation-based approximation analogous to that in Section~\ref{sec:state_space_truncation}, except they provide no approximation guarantees. (By contrast, \citet{intermittent} develops a truncation-based approach with theoretical guarantees for a simpler, related setting where the state uncertainty is exogenous.) \citet{nam2021reinforcement,bellinger2021active} focus on \emph{reinforcement learning} (RL) (as opposed to the planning problem considered here). Specifically, \citet{nam2021reinforcement} focuses on developing RL algorithms for a fixed-horizon setting using the generic POMDP solver POMCP \citep{POUCT}. On the other hand, \citet{bellinger2021active} adapts Q-learning for this setting by utilizing a statistical state estimator to achieve a ``higher costed return” -- for every non-state-sensing action, the subsequent state is simply sampled from the belief distribution. An $\epsilon$-greedy action is then taken based on the sampled state to update the Q-table, without leveraging any structure of the belief distribution while choosing the action. \citet{ATM} proposes a policy improvement heuristic referred to as ATM and devises an RL algorithm to learn this heuristic; we contrast the algorithm proposed here to the ATM heuristic in Section~\ref{sec:heuristic}, and also in our numerical case study in Section~\ref{sec:numerics}. Note that none of the above-mentioned papers focuses on the planning problem in a manner that exploits the specific structure of the MDP and from the standpoint of provable optimality/suboptimality guarantees. A related formulation is considered in~\citet{OracularPOMDPs}, which treats “sensing” as a distinct action and applies a discount factor for its cost at each step a sensing action is taken. Aside from this distinction in the problem formulation, the JIV algorithm proposed in this paper is conceptually similar to the ATM heuristic proposed in \citet{ATM}. A similar framework also appears in \citet{timesense}, but for continuous-time problems with emphasis on learning rather than planning. Finally, another related formulation is analysed in \citet{reisinger2024markov}; here, if the agent decides not to sense the state in any epoch, it is constrained to play the same action as in the previous epoch.

\vspace{-0.2cm}
\section{Problem Formulation}
\label{sec:model}
\vspace{-0.2cm}
In this section, we formally define our MDP formulation with a state sensing cost. We do this by first defining a `standard' discounted cost MDP that serves as our baseline; we subsequently incorporate a state sensing cost, and a protocol for opportunistic state sensing on the part of the agent, into this baseline MDP.

\noindent {\bf Baseline MDP:} Consider an infinite horizon discounted cost MDP $\mathcal{M}(\mathcal{S}, A, \mathcal{T}, \mathcal{C}, \alpha).$ Here, 

    $\bullet$ $\mathcal{S} = \{1, 2, \ldots, |\mathcal{S}|\}$ denotes the (finite) state space,\\
    $\bullet$ $A = \{1,2,\ldots,|A|\}$ denotes the (finite) action space,\\
    $\bullet$ $\mathcal{T}$ denotes the transition function (i.e., $\mathcal{T}(s, a, s')$ denotes the probability of transitioning to state~$s'$ on taking action~$a$ in state~$s$),\\
    $\bullet$ $\mathcal{C}$ denotes the cost function (i.e., $\mathcal{C}(s, a)$ is the cost associated with taking action~$a$ in state~$s$),\\
    $\bullet$ $\alpha \in (0,1)$ denotes the discount factor.

With some abuse of notation, for~$a \in A,$ we use $\T(a)$ and $\C(a)$ to denote, respectively, the $|\mathcal{S}|\times|\mathcal{S}|$ transition probability matrix, and the $|\mathcal{S}|\times 1$ (column) vector of costs, associated with the action~$a.$\footnote{Implicit in this notation is the assumption that any action~$a \in A$ can be taken in any state~$s \in \mathcal{S}.$} Denoting the state at time~$t$ by $X_t,$ and the action at time~$t$ by $A_t,$ there is a well-established theory for characterizing and computing the optimal policy that minimizes the expected discounted cost~\[\mathbb{E}\left[\sum_{t=0}^{\infty} \alpha^t \mathcal{C}(X_t,A_t)\right];\] see~\citet{puterman2014markov} and ~\citet{ross1992applied}. We use~$V^*$ and $Q^*$ to denote, respectively, the optimal value function and the optimal action-value function, corresponding to~$\M.$ As is convention, we treat $V^*$ to be an $|\S| \times 1$ column vector, and $Q^*$ to be a $|\S| \times |A|$ matrix.

\noindent {\bf MDP with state sensing cost:} \hspace{-2pt}We now incorporate a state sensing cost \hspace{-1pt}$k\hspace{-2pt}>\hspace{-2pt}0$ to the above baseline MDP. \hspace{-1pt}The interaction protocol between the agent and environment at each time step \hspace{-1pt}\( t\hspace{-2pt} \geq \hspace{-2pt}0 \) is as follows:

% \begin{itemize}
    $\bullet$  The agent takes action~$A_t$ and commits either to sensing the state at the next step (a \emph{sensing action}) or not (a \emph{blind action}).\\
    $\bullet$ Agent incurs cost~$\C(X_t,A_t),$ and an additional sensing cost~$k$ in case it made a \emph{sensing action}.\\
    $\bullet$ The next state $X_{t+1}$ is drawn according to $\T(X_t,A_t,\cdot)$. If the agent made a \emph{sensing action}, then~$X_{t+1}$ is revealed; if it made a \emph{blind action}, then $X_{t+1}$ remains hidden (saving on the sensing cost $k$).
% \end{itemize}

We assume the agent knows its initial state~$X_0$. If it chooses a blind action at time~$t$, then its next action~$A_{t+1}$ must be taken without precise knowledge of~$X_{t+1}$. The agent’s objective is to minimize expected discounted cost (including the state sensing cost), i.e., \[\mathbb{E}\left[\sum_{t=0}^{\infty} \alpha^t \left( \mathcal{C}(X_t,A_t) + k \indicator{\text{sensing action at }t}\right)\right].\]

We formulate the sequential decision problem as an MDP~$\mathcal{M}_k$ with a countably infinite state space. For clarity, we refer to states in the baseline MDP (elements \hspace{-1pt}of \hspace{-1pt}$\S$)\hspace{-0.5pt} as `root states.' \hspace{-2pt}The \hspace{-1pt}state \hspace{-1pt}space \hspace{-1pt}of \hspace{-1pt}$\mathcal{M}_k$\hspace{-1pt} is
\[\mathcal{S}_{\infty} := \mathcal{S} \cup \left[ \mathcal{S} \times \left( \cup_{j=1}^\infty A^j \right) \right].\]
Here, the state variable corresponds to the most recently sensed (root) state, along with the string of blind actions taken thereafter. Each state~$\tilde{s} \in \S_{\infty}$ is associated with a belief distribution $\mathcal{B}(\st) \in \mathbb{R}^{1 \times |\mathcal{S}|}$ over the set of root states. Specifically, for $\st = (s, a_1, \ldots, a_n),$ where $s \in \S$ and $a_i \in A$ for $1 \leq i \leq n,$ 
\[
\mathcal{B}(\st) = e_s \mathcal{T}(a_1) \mathcal{T}(a_2) \cdots \mathcal{T}(a_n),
\]
where $e_s$ denotes the unit row vector with the $s^{\text{th}}$ entry being one. By convention, for $\st = s \in \S,$ (i.e., right after a sensing action), $\mathcal{B}(\st) = e_s.$ Next, the action space $\A$ for $\M_k$ is defined as 
\[\A = A \times \{ \s, \bl \},\] where the second component of the action captures the decision of whether or not to sense the state at the next time step. Thus $\A$ is finite; we also write $\A = \A_s \cup \A_b,$ where $\A_s = A \times \{ \s\}$ are sensing actions, and $\A_b = A \times \{\bl\}$ blind actions.

The cost function $\mathcal{C}_{\infty}: \mathcal{S}_{\infty} \times \A  \rightarrow \mathbb{R}$ associated with~$\M_k$ is defined as follows:
\begin{align*}
    \mathcal{C}_{\infty}(\st,(a,\s)) &= \mathcal{B}(\st) \mathcal{C}(a) + k\\
    \mathcal{C}_{\infty}(\st,(a,\bl)) &= \mathcal{B}(\st) \mathcal{C}(a)
\end{align*}
Note that the cost has been averaged over the belief distribution over the root states.

Finally, we define the transition probability function for $\mathcal{M}_k$ as $\mathcal{T}_{\infty}: \mathcal{S}_{\infty} \times \mathcal{A} \times \mathcal{S}_{\infty} \rightarrow \mathbb{R}$ as follows:
\begin{align*}
\mathcal{T}_{\infty}(\st_1,(a,\bl),\tilde{s}_2) &= \begin{cases} 1, & \text{for } \tilde{s}_2 = (\st_1,a)\\
0, & \text{otherwise} \end{cases}\\
\mathcal{T}_{\infty}(\st_1,(a,\s),\tilde{s}_2) &= \begin{cases} 0, & \text{for } \tilde{s}_2 \notin \mathcal{S}\\
\mathcal{B}(\st)\mathcal{T}(a) e_{\tilde{s}_2}^T, & \text{for } \tilde{s}_2 \in \mathcal{S}
\end{cases}
\end{align*}
The MDP $\M_k,$ modeling opportunistic state sensing with a sensing cost, is defined using $(\mathcal{S}_{\infty}, \A, \mathcal{T}_{\infty}, \mathcal{C}_{\infty}, \alpha)$. Since the state space is countable and the action space finite, there exists an optimal stationary policy ~\citep{puterman2014markov,ross1992applied}; however, the exact computation of the optimal policy is infeasible due to the infinite state space. In Section~\ref{sec:heuristic}, we propose an algorithm based on selective policy improvement, and in Section~\ref{sec:truncation}, iterative schemes for computing an optimal (or near-optimal) policy via state-space truncation together with a lower bound on the optimal value function, allowing us to quantify the suboptimality gap of any policy.

It is worth noting that, while POMDPs can be reformulated as MDPs over belief states (i.e., belief MDPs), such a transformation leads to an uncountably infinite state space, as the belief space forms a continuous $(n-1)$-simplex. In contrast, our formulation—an MDP with sensing costs—can be viewed as a special case of a POMDP that admits a countable state-space representation. This structural property makes our problem significantly more tractable than general POMDPs.

\vspace{-0.2cm}
\section{Selective Policy Improvement}
\label{sec:heuristic}
\vspace{-0.2cm}

In this section, we introduce the Selective Policy Improvement (SPI) algorithm for the opportunistic state sensing MDP. The algorithm is stated formally as Algorithm~\ref{alg:algorithm_new}. For our notation, let $\V^{\pi} \in \mathbb{R}^{|\S| \times 1}$ denote the value function column vector of policy $\pi$ for the root states. The notation $\pi' \overset{\S}{\succeq} \pi$ indicates that the vector $\V^{\pi'}$ is element-wise less than or equal to $\V^{\pi}$. Finally, $\mathsf{\bf max}\ x$ (respectively, $\mathsf{\bf min}\ x$) for a vector~$x$ denotes its maximum (respectively, minimum) entry.

SPI is initialized with a policy $\pi_{init}$ and hyperparameters $\delta$ and \emph{maxsteps}. It iteratively applies the PolicyUpdate routine (Algorithm~\ref{alg:algorithm_update}) until the improvement in the value function falls below $\delta$. The PolicyUpdate routine improves a reference policy $\pi_{ref}$ as follows:

For each root state~$s,$ it searches for an `improving' sequence of blind actions of length at most \emph{maxsteps}, which lowers the action-value function at~$s,$ relative to the reference policy $\pi_{\text{ref}}$. While the set of all such blind action sequences can be quite large, $\mathsf{PolicyUpdate}$ searches this space \emph{selectively} for computational tractability. Specifically, for a root state $s,$ $\pi'$ considers the candidate blind action sequence  $(a_1, \ldots, a_n)$ only when for each~$i,$ $(a_1, \ldots, a_i)$ is an improvement over $(a_1, \ldots, a_{i-1}),$ followed by the optimal sensing action. This allows for an efficient (and greedy) \hspace{-1pt}search \hspace{-1pt}for \hspace{-1pt}an \hspace{-1pt}improving \hspace{-1pt}blind \hspace{-1pt}action \hspace{-1pt}trajectory.

The preceding check is performed using the following functions: for any vector $\Bar{V} \in \mathbb{R}^{|\S| \times 1}$, define 
\begin{align*}
V_{MS}(\B(\st), \Bar{V}) = \min_{a \in A} \bigl(B(\st) \C(a) + \alpha B(\st) \T(a) \Bar{V} \bigr) + k \\
\pi_{MS}(\B(\st), \Bar{V}) = \arg \min_{a \in A} \bigl(B(\st) \C(a) + \alpha B(\st) \T(a) \Bar{V} \bigr)
\end{align*}
Here, $V_{MS}(\B(\st), \Bar{V})$ denotes the value associated with playing the optimal sensing action under belief $\B(\st)$ with terminal values~$\Bar{V}$ (MS stands for \emph{myopic sensing}). $\pi_{MS}$ denotes the corresponding optimal sensing action. Importantly, note that SPI only performs policy evaluations over the \emph{root states} of $\M_k.$

\begin{algorithm}[tb]
\caption{Selective Policy Improvement (\textbf{SPI})}
\label{alg:algorithm_new}
\textbf{Input}: Initial policy $\pi_{init}$ , \emph{maxsteps}, $\delta$ \\
\textbf{Output}: $\pi' \overset{\S}{\succeq}\pi_{init}$
\begin{algorithmic}[1]
\State $\pi' \gets \pi_{init}$
\State $\pi_{\text{improv}} \gets \textsc{PolicyUpdate}(\pi', \text{\emph{maxsteps}})$
\While{$\mathsf{\bf max}\ (\V^{\pi'} -\V^{\pi_{\text{improv}}}) > \delta$}
    \State $\pi' \gets \pi_{\text{improv}}$
    \State $\pi_{\text{improv}} \gets \textsc{PolicyUpdate}(\pi', \text{\emph{maxsteps}})$
\EndWhile
\end{algorithmic}
\end{algorithm}

\begin{algorithm}[tb]
\caption{$\mathsf{Policy Update}$}
\label{alg:algorithm_update}
\textbf{Input}: $\pi_{ref}, \text{\emph{maxsteps}}$ \\
\textbf{Output}: $\pi_o \overset{\S}{\succeq} \pi_{ref}$
\begin{algorithmic}[1] %[1] enables line numbers
    \State $\pi_o \gets \pi_{ref}$
    \For{$s \in \mathcal{S}$}
        \State $\st \gets s$
        \State $steps \gets 0$
        \State $\pi' \gets \pi_{ref}$
        \State $exploredstates \gets \emptyset$
        \While{$steps \leq \text{\emph{maxsteps}}$}
            \State $exploredstates \gets exploredstates \cup \{\st\}$
            \State \scalebox{0.95} {$V_{blind} \gets \min_{a \in \mathcal{A}} \Big( \mathcal{B}(\st)\mathcal{C}(a) 
            + \alpha V_{MS}(\mathcal{B}(\st) \mathcal{T}(a), \V^{\pi_{ref}}) \Big)$}
            \State \scalebox{0.9} {$a_{blind} \gets \arg\min_{a \in \mathcal{A}} \Big(\mathcal{B}(\st)\mathcal{C}(a) + \alpha V_{MS}(\mathcal{B}(\st)\mathcal{T}(a), \V^{\pi_{ref}})\Big)$}
            \If{$V_{MS}(\mathcal{B}(\st), \V^{\pi_{ref}}) \leq V_{blind}$}
                \State $\pi'(\st) \gets (\pi_{MS}(\mathcal{B}(\st), \V^{\pi_{ref}}(\S)), sense)$
                \State \textbf{break} \hfill \textit{Exit the while loop}
            \EndIf
            \State $\pi'(\st) \gets (a_{blind}, blind)$
            \State $steps \gets steps + 1$
            \State $\st \gets (\st, a_{blind})$
        \EndWhile
        \If{$\V^{\pi'}(s) < \V^{\pi_{ref}}(s)$}
            \For{$state \in exploredstates$}
                \State $\pi_o(state) \gets \pi'(state)$
            \EndFor
        \EndIf
    \EndFor
\end{algorithmic}
\end{algorithm}

Finally, in our numerical case studies, we initialize the SPI algorithm with the {\bf A}lways {\bf S}ense (AS) policy, which selects the optimal sensing action at each belief state. The corresponding value function of the AS policy, 
%can be expressed, 
for $\tilde{s} \in \mathcal{S}_{\infty},$ is given by
\begin{align*}
    V_{AS}(B(\tilde{s})) &= \min_{a \in A} \bigl(B(\st) \C(a) + \alpha B(\st) \T(a) V^* \bigr) + \frac{k}{1-\alpha} \\
    &=\mathsf{\bf min}\ B(\tilde{s}) Q^* + \frac{k}{1-\alpha}.
\end{align*}
Here, with some abuse of notation, we parameterize the value function~$V_{AS}$ by the belief vector of state~$\tilde{s},$ rather than by~$\tilde{s}$ directly. The action corresponding to the AS policy is thus~$(\pi_{AS}(\tilde{s}),\s),$ where \[\pi_{AS}(\tilde{s}) = \mathsf{\bf argmin}\ B(\tilde{s}) Q^*.\] Here, $\mathsf{\bf argmin}\ x$ for a row vector~$x$ denotes the column index of its minimum entry. It is important to note that AS policy~$\pi_{AS}$ agrees with the optimal policy~$\pi^*$ associated with the baseline MDP~$\M$ over root states. In Section~\ref{sec:AS_optimality}, we show that the AS policy is also optimal for~$\M_k$ when the sensing cost~$k$ is small.
\vspace{-0.2cm}
\subsection{Step-by-Step Walkthrough of \texorpdfstring{$\mathsf{Policy\ Update}$}{Policy Update}}
\vspace{-0.2cm}
Starting at every root state $s$, $\mathsf{Policy Update}$ initializes the candidate policy $\pi'$ with $\pi_{ref}$. It then searches for a sequence of actions to update $\pi'$ and improve upon $\pi_{ref}$ at the specific root state $s$. For the given root state, the output policy $\pi_o$ plays the sequence of action that performs better at that root state.

\textbf{Lines 9--10.} For a given belief state $\tilde{s}$ (initialized by the root state $s$), the algorithm identifies the blind action $a_{blind}$ as follows:
For each blind action $(a, blind) \in \mathcal{A}_{b}$, it computes the cumulative cost, which consists of:

$\bullet$ Executing the blind action $(a, blind)$, incurring an immediate cost $\mathcal{B}(\tilde{s}) \mathcal{C}(a)$.\\
$\bullet$ Transitioning to the belief state $(\tilde{s}, a_{blind})$ and obtaining the subsequent return $V_{MS}(\mathcal{B}(\tilde{s})\mathcal{T}(a), V^{\pi_{ref}}_{\mathcal{M}_k}).$

Note that this subsequent cumulative cost of $V_{MS}(\mathcal{B}(\tilde{s})\mathcal{T}(a), V^{\pi_{ref}}_{\mathcal{M}_k})$is obtained by first executing the sensing action given by $\pi_{MS}(\mathcal{B}(\tilde{s}) \mathcal{T}(a), V^{\pi_{ref}}_{\mathcal{M}_k})$ at the belief state $(\tilde{s}, a_{blind})$, and then following the reference policy $\pi_{ref}$ thereafter. The minimum cumulative cost over all blind actions is denoted $V_{blind}$, with the corresponding minimizing action denoted $a_{blind}$.

\textbf{Lines 11--17.} If the cumulative cost of playing the optimal sensing action at $\tilde{s}$ given by $V_{MS}(\mathcal{B}(\tilde{s}), V^{\pi_{ref}}_{\mathcal{M}_k}),$
is lower than $V_{blind}$, the algorithm terminates the search and updates the candidate policy $\pi'$ with the obtained sequence of actions, including the final sensing action. Otherwise (lines 15-17), the process repeats (up to a maximum of $maxsteps$) for the subsequent belief state $(\tilde{s}, a_{blind})$ obtained by executing the action $(a_{blind}, blind)$ on state $s$.

\textbf{Lines 19--23.} The value function $\V^{\pi'}(s)$ of the policy $\pi'$ at $s$, which evaluates the cumulative cost of the candidate action sequence, is compared to that of the reference policy $\pi_{ref}$. The output policy $\pi_o$ then adopts the sequence of actions starting from $s$ that corresponds to the policy with the lower value function.

This process is repeated for all root states, constructing the final output policy $\pi_o$ by aggregating the improved action sequences for each root state where the value of $\pi'$ is lower than that of the reference policy $\pi_{ref}$.

It is instructive to compare SPI with the ATM heuristic proposed in \cite{ATM}. The latter is more restricted in its search for good blind actions; in any belief state~$\tilde{s},$ it seeks to improve upon $\pi_{AS}$ by comparing the actions $(\pi_{AS}(\tilde{s}), \textit{blind})$ and $(\pi_{AS}(\tilde{s}), \textit{sense}).$ For a detailed analysis of the ATM heuristic, we refer the reader to Section~\ref{sec:ATM Analysis}.

\vspace{-0.2cm}
\section{\hspace{-5pt}Solving \hspace{-1pt}\texorpdfstring{$\M_k$}{Mk}\hspace{-1.5pt} with \hspace{-1pt}Provable \hspace{-1pt}Guarantees}
\label{sec:truncation}
\vspace{-0.2cm}
In this section, we establish conditions and methods that yield provable optimality and suboptimality guarantees for any policy $\pi$ of MDP~$\M_k$. We first present a sufficient condition under which always sensing is optimal. We then analyze a truncated version of~$\M_k$ to derive a lower bound on its optimal value function, which in turn provides a computable bound on the suboptimality gap of any policy.

\vspace{-0.2cm}
\subsection{Optimality of always sensing}
\label{sec:AS_optimality}
\vspace{-0.2cm}

The following theorem shows that if the state sensing cost~$k$ is smaller than a certain threshold, then it is optimal to always sense the state.
\begin{theorem}
\label{thm:always_sense}
    If $$k < \alpha\ \mathsf{\bf min} \min_{a_1, a_2 \in A} \left[ \mathcal{T}(a_1)\left(Q^\ast(a_2) - V^\ast\right) \right],$$ 
    then the AS policy 
    defined in~Section~\ref{sec:heuristic} 
    is optimal for the  MDP $\mathcal{M}_{k}.$ 
    %then the policy sensing actions are optimal at every state $\st \in \mathcal{S}_{\infty}$ for the MDP $\mathcal{M}_{k}$
\end{theorem}

The threshold on state sensing cost in Theorem~\ref{thm:always_sense} is strictly positive if and only if, for any action~$a_1$ taken in any root state~$j,$ there does not exist an action~$a_2$ that is optimal in the baseline MDP on all states that lie in the belief support.
\vspace{-0.2cm}
\subsection{Analysis via state space truncation}
\label{sec:state_space_truncation}
\vspace{-0.2cm}
We now consider a class of finite MDPs obtained via state space truncation of~$\M_k.$ Specifically, the truncation is parameterized by~$n \geq 0,$ the maximum number of consecutive blind actions the agent is permitted to take. Formally, the truncated MDP, denoted by~$\M_{k,n}$ is defined as follows. The state space is given by 
\[\mathcal{S}_{n} := \mathcal{S} \cup \left[ \mathcal{S} \times \left( \cup_{j=1}^n A^j \right) \right].\] 
For $n \geq 1$, this yields $\S = \S_0 \subset \S_n \subset \S_{n+1} \subset \S_{\infty}.$ It is convenient to categorize the states of~$\S_n$ into `layers' as follows: Let $\mathcal{L}_0$ be the set of root states (the $0^{\text{th}}$ layer), where the agent knows its current state precisely. Next, we define $\mathcal{L}^j_m$ as the set of `descendants' of the root state $j$ in the $m^{\text{th}}$ layer, i.e., the set of states corresponding to playing $m$ successive blind steps starting from the root state $j$. More formally, 
{\small \[
\mathcal{L}^j_m = \hspace{-1pt} \{\st \in \mathcal{S}_{\infty} \hspace{-2pt} \mid \hspace{-1pt} \st= (j, a_1, \ldots, a_m), \hspace{-3pt} \text{ where } a_1,\ldots,a_m \in A\}.
\]
}Finally, $\mathcal{L}_m$ defines the $m^{\text{th}}$ layer, defined as the union of sets $\mathcal{L}^j_m$ over all root states $j$ in $\mathcal{L}_0$, i.e., 
$\mathcal{L}_m = \bigcup_{j \in \mathcal{L}_0} \mathcal{L}^j_m.$ Note that~$\S_n = \cup_{m=0}^n \mathcal{L}_m.$ Figure~\ref{fig:mdp1} provides an illustration of this layered view of the state space~$\S_n$ for the special case of a two-state ($\S = \{0,1\}$), two-action ($A = \{L,R\}$) baseline MDP.

The action space $\A_{n}$ of $\M_{k,n}$ is simply $\A_{\infty}$, whereas the transition function $\mathcal{T}_{n}: \mathcal{S}_n \times \A_{n} \times \mathcal{S}_n \rightarrow \mathbb{R}$ and the cost function $\mathcal{C}_{n}: \mathcal{S}_n \times \A_{n} \rightarrow \mathbb{R}$ are given by:
\begin{align*}
\mathcal{T}_{n}(\st_1, (a, \cdot), \tilde{s}_2) &= 
\begin{cases} 
\mathcal{T}_{\infty}(\st_1, (a, \cdot), \tilde{s}_2), & \text{if } \st_1 \notin \mathcal{L}_n \\
\mathcal{T}_{\infty}(\st_1, (a, \s), \tilde{s}_2), & \text{otherwise}
\end{cases}
\end{align*}
\begin{align*}
\mathcal{C}_{n}(\st, (a, \cdot)) &= 
\begin{cases} 
\mathcal{C}_{\infty}(\st, (a, \cdot)), & \text{if } \st \notin \mathcal{L}_n \\
\mathcal{C}_{\infty}(\st, (a, \s)), & \text{otherwise}
\end{cases}
\end{align*}
Note that the transition and cost functions in~$\M_{k,n}$ agree with those in~$\M_k,$ except on states at the $n^{\text{th}}$ layer, where state sensing is enforced.

Since~$\M_{k,n}$ is a finite MDP, it admits an exact computation of its optimal policy~$\pi^*_{\M_{k,n}}$ and optimal value function~$V^*_{\M_{k,n}}.$ Of course, the complexity of this computation grows exponentially in~$n,$ so this is only feasible for small values of~$n.$ In the remainder of this section, we relate the solutions of the (finite, and therefore `tractable') truncated MDPs $\{\M_{k,n}\}$ to one another, and to the solution of $\M_k.$ 

Our first result bounds the suboptimality induced by the aforementioned state space truncation.
\begin{theorem}
\label{thm:depththreshold}
$$V^*_{\M_{k,N}}(j) - V^*_{\M_{k}}(j) \le  \frac{\alpha^N k}{1-\alpha} \quad \forall j \in \S,N\ge 0.$$ 
\end{theorem}
Note that using the above result, one can determine a suitable truncation depth~$N$ given a suboptimality tolerance, \emph{without having to solve $\M_{k,N}$ first}. Moreover, solving $\M_{k,N}$ yields a computable lower bound on the optimal value function $V^*_{\M_k}$ of $\M_k$, which can then be used to bound the suboptimality gap of any policy. (A sharper bound, expressed in terms of the solution of $\M_{k,N}$ is provided later in Theorem~\ref{thm:pi_N_optimal}.)

Our next result gives a necessary and sufficient condition for the optimal policy for $\M_{k,n}$ to also be optimal for $\M_{k,n+1}.$ For $s \in \mathcal{S}$ and $a_1,a_2, \ldots, a_m \in A$, define
\begin{multline*}
Z((s,a_1,a_2, \ldots, a_m)) := \mathcal{C}(s,a_1) \\
+ \sum_{i=1}^{m-1} \alpha^i \mathcal{B}((s,a_1,a_2, \ldots, a_i)) \mathcal{C}(s, a_{i+1})
\end{multline*}
as the average cumulative discounted cost incurred in reaching the state $\tilde{s}=(s,a_1,a_2, \ldots, a_m)$ from its root state $s$ (by taking a sequence of blind steps).
\begin{figure}[t]
\centering
\begin{tikzpicture}[level distance=1.2cm, scale=0.8] % Adjusted scale to 0.8
\tikzset{every tree node/.style={minimum width=1.5em,draw,circle},
         blank/.style={draw=none},
         edge from parent/.style=
         {draw,edge from parent path={(\tikzparentnode) -- (\tikzchildnode)}},
}
\Tree
[.\node{$0$}; 
  [.{$0L$}
    [.$0LL$ ]
    [.$0LR$ ]
  ]
  [.$0R$
    [.$0RL$ ]
    [.$0RR$ ]
  ]
]
\end{tikzpicture}
\begin{tikzpicture}[level distance=1.2cm, scale=0.8] % Adjusted scale to 0.8
\tikzset{every tree node/.style={minimum width=1.5em,draw,circle},
         blank/.style={draw=none},
         edge from parent/.style=
         {draw,edge from parent path={(\tikzparentnode) -- (\tikzchildnode)}},
}
\Tree
[.\node{$1$}; 
  [.{$1L$}
    [.$1LL$ ]
    [.$1LR$ ]
  ]
  [.$1R$
    [.$1RL$ ]
    [.$1RR$ ]
  ]
]
\end{tikzpicture}
\caption{\footnotesize \textbf{State space of $\M_{k,2}$} for a 2-state 2-action baseline MDP~$\M$}
\label{fig:mdp1}
\end{figure}
\begin{lemma}
\label{thm:one-step}
Fix $N \geq 0.$
% shortened
$V^*_{\M_{k,N}}(j) = V^*_{\M_{k,N+1}}(j)$ for all root states $j \in \S$ if and only if 
%he ``explored states" $s_{exp}$ corresponding to every root state of $\mathcal{M}_{k,N}$ remains unchanged for $\mathcal{M}_{k,N+1}$ if and only if Inequality \eqref{next} holds for every root state $j$
\begin{align*}
Z(i) + \alpha^{N+1} & \min_{a} \Bigl(\mathcal{B}(i)\mathcal{C}(a) + k + \alpha \mathcal{B}(i)\mathcal{T}(a)V^*_{\M_{k,N}} \Bigr) \\
& \quad \quad \geq V^*_{\M_{k,N}}(j) \quad \forall j \in \S, i \in \mathcal{L}^j_{N+1}.
%G_{N+1}(j,i) \ge V^*_{\M_{k,N}}(j) \quad
%\forall i \in \mathcal{L}^j_{N+1} \label{next}
\end{align*}
\end{lemma}
Note that $V^*_{\M_{k,N}}(j) = V^*_{\M_{k,N+1}}(j)$ for all $j \in \S$ implies that the optimal stationary policy $\pi^*_{M_{k,N}}$ for $\M_{k,N}$ is also optimal for $\M_{k,N+1}$ when starting in a root state (i.e., with knowledge of the starting state). However, this condition \emph{does not} imply that $\pi^*_{\M_{k,N}}$ is optimal for $\M_{k}$ when starting in a root state, as shown by a counterexample in Appendix~\ref{sec:counterexample}. One needs a stricter condition on $\M_{k,N}$ to conclude that $\pi^*_{\M_{k,N}}$ is optimal for $\M_k;$ this is the focus of our next result. 

Define 
\begin{align*}
    V_{AS;0}(\tilde{s}) :=\mathsf{\bf min}\ B(\tilde{s}) Q^*.
\end{align*}
 This is simply the value function corresponding to the AS policy introduced in Section~\ref{sec:heuristic}, assuming zero sensing cost. This means $V_{AS;0}$ provides a \emph{lower bound} on the optimal value function for $\M_k.$

\begin{theorem}
\label{thm:pi_N_optimal}
Fix $N \geq 0.$ If 
\begin{equation}
Z(i) + \alpha^{N+1} V_{AS;0}(i) \ge V^*_{\M_{k,N}}(j) \quad \forall j \in \S,\ i \in \mathcal{L}^j_{N+1},
\label{ultimate}
\end{equation}
then the optimal stationary policy $\pi^*_{\M_{k,N}}$ of $\M_{k,N}$ is optimal for $\M_k$ when starting at any root state.
If~\eqref{ultimate} does not hold, 
\begin{align}
V^*_{\M_{k,N}}(s) - V^*_{\M_{k}}(s) \le \epsilon_N \quad \forall s\in \S,
\label{eqn:suboptimalitybound}
\end{align}where 
$$\epsilon_N := \max_{j \in S} \left[V^*_{\M_{k,N}}(j) - \min_{i \in \mathcal{L}^j_{N+1}} \left(Z(i) + \alpha^{N+1} V_{AS;0}(i) \right)\right].$$
\end{theorem}
Theorem~\ref{thm:pi_N_optimal} provides a sufficient condition~\eqref{ultimate} for the optimal policy for $\M_{k,n}$ to also be optimal for $\M_k$ (assuming the starting state is a root state). Even if this condition is violated, Theorem~\ref{thm:pi_N_optimal} provides a computable upper bound on the suboptimality of the policy $\pi^*_{\M_{k,N}}$ on $\M_k.$ Thus, Theorem~\ref{thm:pi_N_optimal} suggests a recipe for computing an optimal/near-optimal policy for $\M_k$: Iteratively solve $\M_{k,N}$ for increasing~$N,$ until either (i) the condition~\eqref{thm:pi_N_optimal} is satisfied, in which case the optimal policy just computed is also optimal for $\M_k,$ or (ii) the suboptimality bound $\epsilon_N$ is acceptably small. Moreover, analogous to Theorem~\ref{thm:depththreshold}, one can also derive a (stronger) lower bound on the optimal value function $V^*_{\M_k}$ of $\M_k$.

As shown in Appendix~\ref{sec:extension}, even if \eqref{ultimate} is satisfied only at certain root states, one does not need to explore depths~$N+1$ and beyond at these root states. Furthermore, Lemma \ref{thm:epsilon} establishes that the suboptimality bound $\epsilon_N$ decreases monotonically in $N$.
\vspace{-0.3cm}
\section{Numerical Case Studies}
\label{sec:numerics}
\vspace{-0.2cm}
In this section, we present numerical experiments that validate and complement the analytical results from prior sections. We also benchmark the proposed approaches against the state of the art.

{\bf 2-State 2-Action example:} We consider a two-state, two-action baseline MDP, shown in Figure~\ref{fig:example2}, and evaluate it under sensing costs (a) $k=0.01$ and (b) $k=0.25$.

$\bullet$ Applying Theorem~\ref{thm:always_sense} yields a sensing cost threshold of $0.05$, making always sensing optimal in case$(a).$ Thus, the optimal policy for state $0$ is $(R, \s)$, and for $1$ is $(B, \s)$.\\
$\bullet$ For $k=0.25$, the optimal policy for both root states remains unchanged after $N\ge 2$.  Indeed, the condition of Lemma~\ref{thm:one-step} is satisfied at $N=2$. However, the condition of Theorem~\ref{thm:pi_N_optimal} is satisfied for $N=4$ which suggests that the optimal policy should remain unchanged for $N\ge 4;$ see Figure~\ref{fig:II}. \\
(The stronger lower bound on $V^*_{\M_k}$ used in Figure~\ref{fig:II} is detailed in Appendix~\ref{sec:extension}.)

% \begin{figure}[t]
% \centering
% \begin{tikzpicture}[state/.style={draw, circle, minimum size=0.75cm},>=Stealth,node distance=3cm]

% % Nodes
% \node[state] (0) {0};
% \node[state, right=of 0] (1) {1};

% % Edges
% \draw[->, blue] (0) to[bend right=20] node[midway, below]{$0.7$, $1$} (1);
% \draw[->, red] (0) to[bend right=70] node[midway, below]{$0.3$, $0$} (1);
% \draw[->, blue] (1) to[bend left=-20] node[midway, above]{$0.9$, $0$} (0);
% \draw[->, red] (1) to[bend left=-70] node[midway, above]{$0.2$, $1$} (0);
% \draw[->, red] (0) to [out=-135,in=135,loop,looseness=4.8] node[below=1.3cm of 0] {$0.7$, $0$} (0);
% \draw[->, blue] (0) to [out=-120,in=120,loop,looseness=10.8] node[above=1.3cm of 0] {$0.3$, $1$} (0);
% \draw[->, red] (1) to [out=45,in=-45,loop,looseness=4.8] node[below=1.3cm of 1] {$0.8$, $1$} (1);
% \draw[->, blue] (1) to [out=60,in=-60,loop,looseness=10.8] node[above=1.3cm of 1] {$0.1$, $0$} (1);

% \end{tikzpicture}
% \caption{A two-state two-action baseline MDP with actions $\{Red, Blue\}$ and and $\alpha = 0.5$}
% \label{fig:example2}
% \end{figure}
\begin{figure}[t]
\centering
\begin{tikzpicture}[state/.style={draw, circle, minimum size=0.75cm},>=Stealth,node distance=3cm]

% Define professional red/blue palette
\definecolor{neublue}{RGB}{33,113,181}   % professional deep blue
\definecolor{neured}{RGB}{200,33,33}     % professional deep red

% Nodes
\node[state] (0) {0};
\node[state, right=of 0] (1) {1};

% Edges
\draw[->, neublue, thick] (0) to[bend right=20] node[midway, below]{$0.7$, $1$} (1);
\draw[->, neured, thick, dashed] (0) to[bend right=70] node[midway, below]{$0.3$, $0$} (1);

\draw[->, neublue, thick] (1) to[bend left=-20] node[midway, above]{$0.9$, $0$} (0);
\draw[->, neured, thick, dashed] (1) to[bend left=-70] node[midway, above]{$0.2$, $1$} (0);

\draw[->, neured, thick, dashed] (0) to [out=-135,in=135,loop,looseness=4.8] node[below=1.3cm of 0] {$0.7$, $0$} (0);
\draw[->, neublue, thick] (0) to [out=-120,in=120,loop,looseness=10.8] node[above=1.3cm of 0] {$0.3$, $1$} (0);

\draw[->, neured, thick, dashed] (1) to [out=45,in=-45,loop,looseness=4.8] node[below=1.3cm of 1] {$0.8$, $1$} (1);
\draw[->, neublue, thick] (1) to [out=60,in=-60,loop,looseness=10.8] node[above=1.3cm of 1] {$0.1$, $0$} (1);

\end{tikzpicture}
\caption{\footnotesize \textbf{A two-state two-action baseline MDP} with actions $\{Red, Blue\}$ and $\alpha = 0.5$.}
\label{fig:example2}
\end{figure}
\begin{figure}[t]
    \centering  \includegraphics[width=\linewidth]{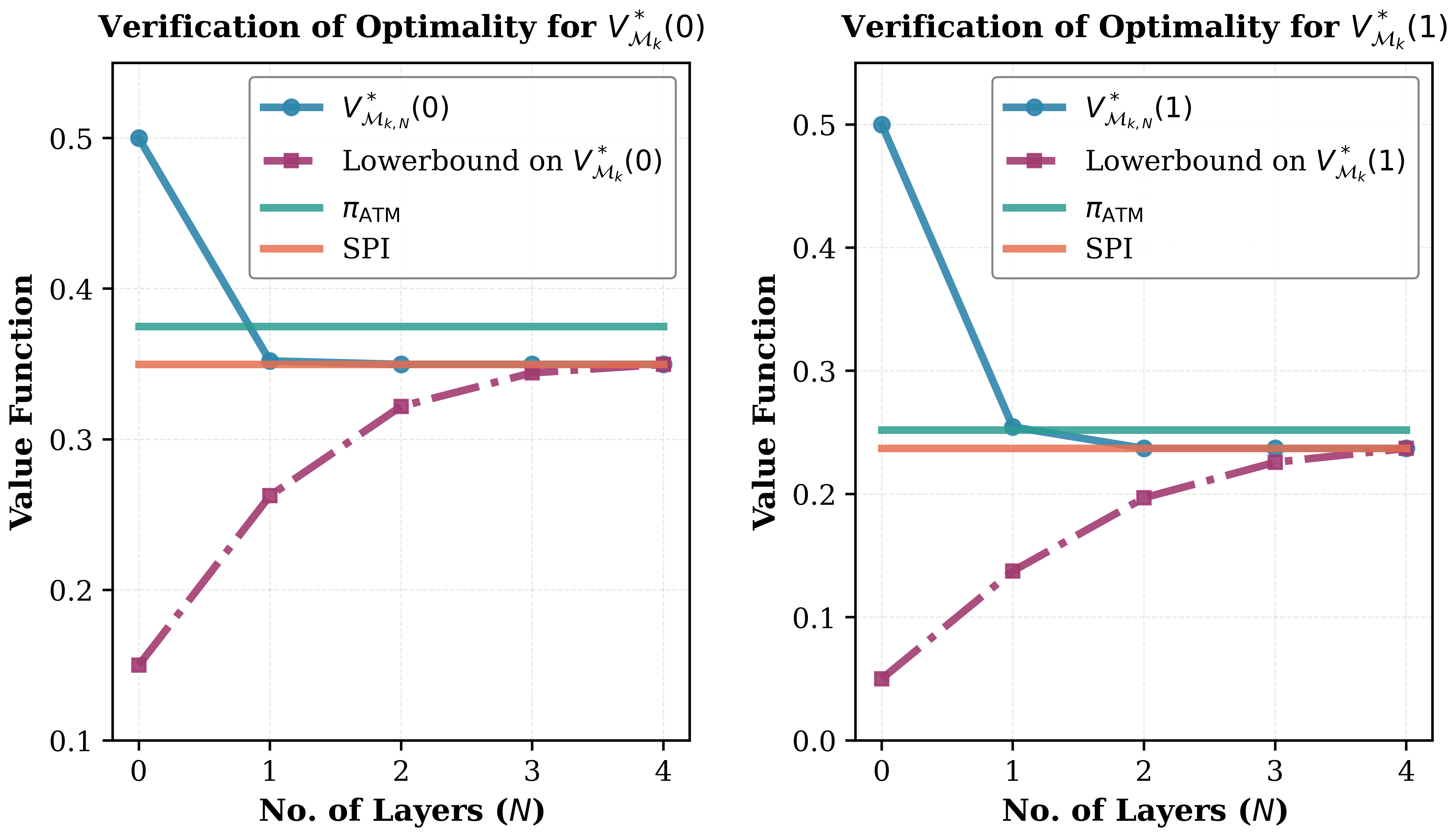}
    \caption{\footnotesize \textbf{Applying Thm~\ref{thm:pi_N_optimal}} on MDP in Fig~\ref{fig:example2}; $k = 0.25$}
    \label{fig:II}
\end{figure}

\textbf{Experimental Setup.} We benchmark the proposed SPI algorithm against the ATM heuristic \citep{ATM}, alongside widely used general-purpose POMDP planning algorithms, including SARSOP \citep{SARSOP}, Fast Informed Bound (FIB) \citep{FIB}, using the \texttt{POMDPs.jl} framework \citep{juliapomdps}. Experiments are conducted across the ICU-Sepsis benchmark \citep{icu} and gridworld-based environments from Gymnasium \citep{gymnasium}, including Taxi and Frozen Lake. Algorithms such as Incremental Pruning \citep{cassandra1997incremental} and PBVI \citep{PBVI} exhibited prohibitive runtimes, failing to produce effective policies even for small state spaces, while other methods like QMDP \citep{QMDP} were consistently outperformed by our approach. Even popular online planning algorithms like PO-UCT \citep{POUCT} required significantly higher planning times to generate competitive policies. 

These POMDP planners were initialized with a belief distribution corresponding to the starting state distribution, rather than a uniform distribution over all states. The experimental design, along with an analysis of POMCP’s performance, is provided in  Appendix~\ref{POUCT}. For SARSOP, the reward and sensing costs were scaled by 10\textsuperscript{3} during policy computation to match performance on Frozen Lake.
 
{\bf ICU-Sepsis ($|\S| = 716 \  \& \ |A|= 25$) :} \citet{icu} is a tabular MDP modeling personalized care for sepsis patients in the ICU, derived from electronic health record (EHR) data. The \emph{state space} $\mathcal{S}$ models a patient's health as a vector of clustered features (demographics, vital signs, body fluid levels) discretized into 716 states. The \emph{action space} consists of discrete combinations of medication types and dosages. The agent receives 0 reward for non-terminal steps and +1 for survival ($\alpha = 0.99$).\\
{\bf Frozen Lake ($|\S| = 16/64 \  \& \ |A|= 4$) :}
The agent navigates across a frozen lake and receives a reward of $+1$ upon reaching the goal state ($\alpha = 0.9$). In the slippery case, actions succeed with probability $\frac{1}{3}$, otherwise moving perpendicularly with equal probability $\frac{1}{3}$. We evaluate default 4x4, 8x8, and a custom, challenging 4x4 Frozen Lake grid (see Table~\ref{table:FrozenLakeInstance}). For Frozen Lake, we report expected reward over the initial state \textbf{S} and maximum computation times across sensing costs.\\
{\bf Stochastic Taxi ($|\S| = 500 \  \& \ |A|= 6$) :} The agent navigates a 5x5 grid to pick up and drop off passengers, receiving rewards of +20 for delivery, -10 for illegal “pickup” and “drop-off” actions, and -1 per step unless another reward is triggered ($\alpha=0.95$)~\citep{taxi}.  Each of the four navigation actions moves the agent in the intended direction with a probability of 0.8, and in one of the two perpendicular directions with an equal probability of 0.1. The initial states are sampled from 300 valid states.
\begin{table*}[t]
\footnotesize
\centering
\begin{minipage}{0.48\textwidth}
\centering
\resizebox{\columnwidth}{!}{
    \begin{tabular}{@{}lccccc@{}}
        \toprule
        \textbf{Environment} & \multicolumn{4}{c}{\textbf{Sensing Costs}} \\
        \cmidrule{2-5}
        & \textbf{0.005} & \textbf{0.01} & \textbf{0.05} & \textbf{0.1} \\
        \midrule
        \textbf{ICU-Sepsis} & & & & \\
         \hspace{0.2cm} \multirow{2}{*}{\hspace{-0.21cm} $\pi_{\text{ATM}}$} 
        \scriptsize \hspace{0.03cm} \textbf{Value} & 0.740 & 0.715 & 0.714 & 0.714 \\
        \hspace{1.15cm} \scriptsize \textbf{Time} (s)  & 285 & 782 & 833 & 916  \\
        \scalebox{0.92}{SARSOP (3000s)} 
         & 0.741 & 0.741 & 0.741 & 0.740 \\
         \rowcolor{myblue}
        \multirow{2}{*}{\parbox{1cm}{\centering SPI \\ \scriptsize (2 iter)}}
        \scriptsize \hspace{0.01cm} \textbf{Value} & \textbf{0.765} & \textbf{0.747} & \textbf{0.742} & \textbf{0.745} \\
        \rowcolor{myblue}
        \multirow{-2}{*}{\parbox{1cm}{\centering SPI \\ \scriptsize (2 iter)}}
        \hspace{0.01cm} \scriptsize \textbf{Time} (s)  & 23 & 1252 & 2205 & 2757 \\
        \midrule
        & \textbf{0.1} & \textbf{0.5} & \textbf{1} & \textbf{5} \\
        \cmidrule{2-5}
        \textbf{Stochastic Taxi} & & & & \\
        \hspace{0.2cm} \multirow{2}{*}{$\pi_{\text{ATM}}$} 
        \scriptsize \hspace{0.025cm} \textbf{Value} & 0.908 & -0.359 & -2.761 & -19.664 \\
        \hspace{1.25cm} \scriptsize \textbf{Time} (s)  & 2 & 1.7 & 1.8 & 38.4\\
        \scalebox{0.92}{SARSOP (100s)} & -22 & -1.2 & -2.705 & -20  \\
         \rowcolor{myblue}
        \hspace{0.2cm} \multirow{2}{*}{SPI} 
        \scriptsize \hspace{0.28cm} \textbf{Value} & \textbf{0.911} & \textbf{-0.306} & \textbf{-1.598} & \textbf{-9.778}  \\
        \rowcolor{myblue}
        \hspace{0.2cm} \multirow{-2}{*}{SPI}
        \hspace{0.28cm} \scriptsize \textbf{Time} (s)  & 25.2 & 19.7 & 17.8 & 117.8 \\
        \bottomrule
    \end{tabular}
}
\end{minipage}
\hfill
\begin{minipage}{0.5\textwidth}
\centering
\resizebox{\textwidth}{!}{
\begin{tabular}{@{}lccccc@{}}
\toprule
\textbf{Scenario} & \multicolumn{4}{c}{\textbf{Value (Rewards) for Sensing Costs}} & \textbf{Time (s)} \\
\cmidrule(r){2-5}
\textbf{Frozen Lake} & \textbf{0.001} & \textbf{0.005} & \textbf{0.01} & \textbf{0.05} & \\
\midrule
\textbf{4x4 (Default)} & & & & & \\
\quad $\pi_{\text{ATM}}$ & \textbf{62.42} & 36.52 & 6.72 & 16.57 & 0.04 \\
\quad $V^*_{\mathcal{M}_{k,3}}$ & \textbf{62.42} & \textbf{36.53} & 20.47 & -28.75 & 11.5 \\
\quad SARSOP & \textbf{62.42} & \textbf{36.53} & 20.79 & \textbf{23.08} & 1.7 \\
\quad FIB (1000 iter) & \textbf{62.42} & 31.43 & \textbf{23.08} & \textbf{23.08} & 8.3 \\
\rowcolor{myblue}
\quad SPI & \textbf{62.42} & \textbf{36.53} & 20.99 & \textbf{23.08} & 0.4 \\
\midrule
\textbf{4x4 (Hard)} & & & & & \\
\quad $\pi_{\text{ATM}}$ & 8.41 & 0 & 0 & 0 & 0.04 \\
\quad $V^*_{\mathcal{M}_{k,3}}$ & 8.92 & 1.36 & -5.75 & -36.75 & 12 \\
\quad SARSOP & \textbf{8.95} & 3.66 & 1.34 & \textbf{1.44} & 1.6 \\
\quad FIB (1000 iter) & 4.44 & 0 & 0 & 0 & 9.5 \\
\rowcolor{myblue}
\quad SPI & \textbf{8.95} & \textbf{3.69} & \textbf{1.47} & 1.35 & 0.4 \\
\midrule
\textbf{8x8} & & & & & \\
\quad $\pi_{\text{ATM}}$ & 3.29 & 3.29 & 3.29 & 3.29 & 0.25 \\
\quad $V^*_{\mathcal{M}_{k,3}}$ & 2.72 & -4.943 & -13.64 & -79.09 & 205 \\
\quad SARSOP & 3.36 & \textbf{3.36} & \textbf{3.36} & \textbf{3.36} & 2 \\
\quad FIB (20 iter) & 3.29 & 3.29 & 3.29 & 3.29 & 475 \\
\rowcolor{myblue}
\quad SPI & \textbf{3.53} & 3.33 & 3.33 & 3.33 & 3 \\
\bottomrule
\end{tabular}
}
\end{minipage}
\caption{\footnotesize \textbf{Performance comparison across different scenarios} with varying sensing costs; rewards scaled by $10^3$ for Frozen Lake. (Unless the no.\ of $\mathsf{PolicyUpdate}$ iter in SPI are specified, $\delta=10^{-6}$.)}
\label{tab:combined-performance}
\end{table*}

\textbf{Benchmarking Results.} Table~\ref{tab:combined-performance} shows that SPI  consistently outperforms $\pi_{\text{ATM}}$ and FIB, and achieves the best performance on ICU-Sepsis and Taxi across all sensing costs (FIB could not scale to these domains). Results from the custom hard setup further indicate that $\pi_{\text{ATM}}$ and FIB may fail to find effective policies even in small state spaces when the planning problem is challenging,  underscoring SPI’s robustness.

We emphasize that, although SARSOP exhibited performance comparable to SPI for Frozen-Lake, achieving this required scaling the reward and sensing values. Without such scaling, SARSOP performed poorly, even relative to $\pi_{\text{ATM}}$ and FIB. Furthermore, in Stochastic Taxi, SARSOP consistently underperformed, and scaling the values did not improve its performance. Finally, we note that SARSOP's performance is highly sensitive to several hyperparameters that are difficult to interpret and tune. In contrast, SPI relies on only two hyperparameters, both of which are intuitive and easily adjustable. Moreover, SARSOP requires the starting state distribution to compute its policy and can degrade significantly under mismatch, whereas SPI operates without requiring such information.

Finally, our analysis reveals significant limitations in existing RL algorithms for this setting, notably the Observe-before-planning \citep{nam2021reinforcement} and Dyna-ATMQ \citep{ATM}. Observe-before-Planning leverages the ACNO-MDP structure during exploration but resorts to the generic POMDP solver POMCP (PO-UCT) for planning, failing to utilize the specific ACNO-MDP structure during exploitation. Furthermore, PO-UCT performs significantly worse than SPI, especially in scenarios with sparse rewards. Similarly, Dyna-ATMQ employs a less effective planner $\pi_{\text{ATM}}$, which struggles even in small, challenging state spaces. We attribute these inefficiencies to the absence of a strong theoretical foundation for ACNO-MDPs, a gap our paper addresses. By reformulating this setting using an ``expanded state-space setup," we provide novel insights that enable the design of effective planning algorithms like SPI.
\vspace{-0.2cm}
\section{Discussion and Conclusion}
\vspace{-0.2cm}
In this paper, we have analysed a class of MDPs with a state sensing cost. Here, the agent must, in a history-dependent, opportunistic manner, determine when to sense the state of the system. While these MDPs are intractable under generic planning algorithms, we exploit the special structure of these sensing-cost MDPs to devise clever algorithms and truncation approaches with provable optimality/suboptimality guarantees. Our main contributions are summarized as follows:

$\bullet$ We reformulate opportunistic state sensing in an expanded state-space framework and propose a novel planning algorithm, SPI. SPI outperforms state-of-the-art POMDP solvers and specialized solvers for this setting (e.g., $\pi_{\text{ATM}} $) across diverse domains for all ranges of sensing costs within reasonable compute time (Table~\ref{tab:combined-performance}). Moreover, it scales to large-state-space MDPs and naturally extends to continuous state spaces using function approximation (Appendix~\ref{sec:adapt}).\\
$\bullet$ Using truncated MDP analysis, we derive lower bounds on the optimal value function, which enable explicit computation of suboptimality gaps for any policy (Theorems~\ref{thm:pi_N_optimal} and \ref{thm:depththreshold}).We establish a quantitative threshold on the sensing cost (Theorem~\ref{thm:always_sense}) below which always sensing is provably optimal.\\
To the best of our knowledge, this work presents the first set of results establishing suboptimality guarantees for policies and explicit cost thresholds in this setting.

In practice, sensing costs are often context-dependent, shaped by factors such as resource constraints or belief uncertainty (e.g., in medical decision-making or robotics). Our framework is readily extendable to non-uniform sensing costs: both SPI and the truncated MDP approach, along with the theoretical results, apply beyond the constant-$k$ setting (see Appendix~\ref{sec:non-uniform}).

At a high level, this work is related to the vast recent literature on Age of Information (AoI), where the goal is to allocate resources or incur costs so as to minimize the age (a.k.a., staleness) of the state information; see \cite{yates2021age} for a survey. However, the AoI literature does not, as such, consider the \emph{control} aspect where the goal of state estimation is actually to influence the state evolution favourably. Additionally, the AoI literature employs a universal (state-independent) age/staleness penalty; in practice, one would expect that the agent would be more tolerant of delayed state information in certain states than others. The present formulation seeks to formally capture this trade-off.

\newpage
\bibliography{ICML.bib}
\bibliographystyle{plainnat}

\clearpage
\appendix
\thispagestyle{empty}

% Supplementary material: To improve readability, you must use a single-column format for the supplementary material.
\onecolumn
\aistatstitle{Appendix}
\addcontentsline{toc}{section}{Appendix}
\appendixcontents
\section{Proofs}
\label{sec:proofs}
% \customtoc
\subsection{Proof of Theorem~\ref{thm:always_sense}}
\label{proof:sense_thr}
\begin{proof}
We define $\pi$ to be the AS policy for the MDP $\mathcal{M}_k$ and hence for each $\tilde{s} \in \mathcal{S}_{\infty}, \pi(\tilde{s}) = (\pi_{AS}(\tilde{s}), sense)$. It is important to note that AS policy $\pi_{AS}$ takes the optimal sensing action in each belief state and agrees with the optimal policy $\pi^*$ associated with the baseline MDP $\mathcal{M}$ over root states. Therefore, we have:

\begin{align*}
V^\pi_{\mathcal{M}_k}(\tilde{s}) &= \min_{a \in A} \left( B(\tilde{s}) \mathcal{C}(a)+ \alpha B(\tilde{s}) \mathcal{T}(a) V^* \right) + \frac{k}{1-\alpha} \\
&=\min_{a \in A} \left( B(\tilde{s}) \mathcal{C}(a) + \alpha B(\tilde{s}) \mathcal{T}(a) V_{\mathcal{M}_k}^\pi \right) + k \quad (\text{Since } V_{\mathcal{M}_k}^\pi(s) = V^*(s) + \frac{k}{1-\alpha} \quad \forall s \in \mathcal{S})\\
&= \min_{a \in A} Q_{\mathcal{M}_k}^\pi(\tilde{s}, (a, sense))
\end{align*}

Hence, policy $\pi$ cannot be improved by any sensing action for any of the states, i.e.,
\begin{align*}
    Q^{\pi}_{\M_k}(\st, (a, \s)) \ge V^{\pi}_{\M_k}(\st) \quad \forall \st \in \mathcal{S}_{\infty}, \ a \in A.
\end{align*}

\begin{equation}
\begin{aligned}
    Q^{\pi}_{\M_k}(\st, (a_1, \bl)) &\geq  Q^{\pi}_{\M_k}(\st, (a_1, \s)) \quad \forall (\st, a_1) \in \mathcal{S}_{\infty} \times A.
\end{aligned}
\label{sensing}
\end{equation}
We claim that ~\eqref{sensing} is a sufficient condition for $\pi$ to be optimal: If for every state $\tilde{s} \in \mathcal{S}_{\infty}$ and every action $a \in A$, the action-value function for the blind action $(a, blind)$ is greater than or equal to that for the corresponding sensing action $(a, sense)$, i.e., $Q^\pi_{\mathcal{M}_k}(\tilde{s}, (a, blind)) \ge Q^\pi_{\mathcal{M}_k}(\tilde{s}, (a, sense))$, then $\pi$ is optimal. This follows from the initial claim that $Q^\pi_{\mathcal{M}_k}(\tilde{s}, (a, sense)) \ge V^\pi_{\mathcal{M}_k}(\tilde{s})$ holds (indicating that $\pi$ cannot be improved by any sensing action), combined with the fact that a policy satisfying ~\eqref{sensing} cannot be improved by taking blind actions for any states during the Policy Improvement Step.

Criterion \eqref{sensing} holds for a state-action pair $(\st, a_1)$ if and only if
\begin{multline*}
    \mathcal{B}(\st)\mathcal{C}(a_1) + \alpha \left( k + \mathcal{B}(\st)\mathcal{T}(a_1)\mathcal{C}(a_2) \right)
     + \alpha^2 \mathcal{B}(\st)\mathcal{T}(a_1)\mathcal{T}(a_2)V^{\pi}_{\M_k}
    \geq \mathcal{B}(\st)\mathcal{C}(a_1) + k + \alpha \mathcal{B}(\st)\mathcal{T}(a_1)V^{\pi}_{\M_k},
\end{multline*}
where $a_2$ is the action taken according to $\pi$ at the state reached by taking action $a_1$ at state $\st$. Moreover, for any root state $s \in \mathcal{S}$, we have
$$
V^{\pi}_{\M_k}(s) = V_{AS}(\mathcal{B}(s)) = V^\ast(s) + \frac{k}{1 - \alpha}.
$$
Substituting this and rearranging, we get
\begin{align*}
    \frac{k}{1 - \alpha} + \alpha \mathcal{B}(\st)\mathcal{T}(a_1)\mathcal{T}(a_2)V^\ast
    \geq \frac{k}{\alpha (1 - \alpha)} + \mathcal{B}(\st)\mathcal{T}(a_1)V^\ast - \mathcal{B}(\st)\mathcal{T}(a_1)\mathcal{C}(a_2).
\end{align*}
Simplifying, we get
\begin{align*}
    \mathcal{B}(\st)\mathcal{T}(a_1)\mathcal{C}(a_2) - \mathcal{B}(\st)\mathcal{T}(a_1)V^\ast
    + \alpha \mathcal{B}(\st)\mathcal{T}(a_1)\mathcal{T}(a_2)V^\ast 
    \geq \frac{k}{\alpha}.
\end{align*}
Further simplifying, we obtain
$$
\mathcal{B}(\st)\mathcal{T}(a_1)(\mathcal{C}(a_2) + \alpha \mathcal{T}(a_2)V^\ast - V^\ast) \geq \frac{k}{\alpha},
$$
which implies
\begin{equation}
\frac{k}{\alpha} \leq \mathcal{B}(\st)\mathcal{T}(a_1)(Q^\ast(a_2) - V^\ast).
\label{mid}
\end{equation}
If condition~\eqref{mid1} holds, then \eqref{mid} is satisfied for all $(\st, a_1) \in \mathcal{S}_{\infty} \times A$, and hence the AS policy $\pi$ is optimal for $\M_k$.
\begin{equation}
k < \alpha\ \mathsf{\bf min} \min_{a_1, a_2 \in A} \left[ \mathcal{T}(a_1)\left(Q^\ast(a_2) - V^\ast\right) \right].
\label{mid1}
\end{equation}
\end{proof}
\subsection{Proof of Theorem~\ref{thm:depththreshold}}
\label{proof:depth}
\begin{proof}
Consider any stationary policy $\pi$ of $\mathcal{M}_k$ and define the set of root states $G$, such that starting from any state $j \in G$ and following $\pi$, we play at least $N+1$ consecutive blind steps. Also define $\Bar{G}:=\mathcal{S}\setminus G$. 

First, we show that
\begin{align}
& \Vn^{*}(i)-\V^{\pi}(i) \le \sum_{j \in \mathcal{L}_0} p_{ij}  \bigl(\Vn^{*}(j)-\V^{\pi}(j)\bigr) \nonumber \\ &\quad \quad \forall i \in \Bar{G}, \label{eq:Mkn_suboptimality_1}   
\end{align}
where $p_{ij}$ denotes the probability of landing in root state $j \in \mathcal{S}$ after taking the first sensing step when starting from~$i$ and following~$\pi$.

Next, we show that
\begin{align}
    \Vn^*(i)-\V^{\pi}(i) &\le \frac{\alpha^N k}{1-\alpha} , \quad \forall i \in G
    \label{eq:Mkn_suboptimality_2}   
\end{align}

Finally, define $G' = \{s \mid s \in \Bar{G} \text{ and } s \not\rightarrow i, \forall i \in G\}$, i.e., starting from any state $s \in G',$ we never reach a state in $G$ under policy $\pi.$ It is easy to see that 
\begin{align}
     \Vn^{*}(i)-\V^{\pi}(i) &= 0 , \quad \forall i \in G'.
\label{eq:Mkn_suboptimality_3}
\end{align}

We now show how the the statement of the lemma follows from~\eqref{eq:Mkn_suboptimality_1}--\eqref{eq:Mkn_suboptimality_3}. Treat~$f(i):=\Vn^*(i)-\V^{\pi}(i)$ as the reward in state~$i$ corresponding to a Markov chain over~$\mathcal{S}$ with transition probabilities~$\{p_{ij}\},$ the states in $G \cup G'$ being absorbing states. Note that \eqref{eq:Mkn_suboptimality_1} implies that starting in any non-absorbing state, the average reward increases with time; moreover, eventual absorption is guaranteed with probability~1. Since the reward on absorbing states is at most~$\frac{\alpha^N k}{1-\alpha}$ (see~\eqref{eq:Mkn_suboptimality_2} and ~\eqref{eq:Mkn_suboptimality_3}), it follows that $$f(i)=\Vn^*(i)-\V^{\pi}(i) \leq \frac{\alpha^N k}{1-\alpha}\quad \forall \ i \in \mathcal{S}.$$ This implies the statement of the lemma, taking~$\pi$ to be an optimal policy under~$\M_k.$ It now remains to prove~\eqref{eq:Mkn_suboptimality_1} and~\eqref{eq:Mkn_suboptimality_2}.
 
%The set of equations above represents the dynamic programming equations for a Markov Process with a state space $\mathcal{S}$ and zero transition costs, featuring state-dependent discounting factors. Here, $\Vn^{*}(i)-\V^{\pi}(i)$ denotes the value function for the state $i$, and $G \cup G'$ denotes the set of absorbing states. Also, note that any non-absorbing state reaches an absorbing state a.s. (hitting time is finite). Since $a_i \ge 1$ for all $i \in G$, we can conclude that $\Vn^{*}(i)-\V^{\pi}(i) \leq \frac{\alpha^N k}{1-\alpha}$ for all $i \in \Bar{G}\setminus G'$ (for all non-absorbing states).

To prove~\eqref{eq:Mkn_suboptimality_1}, consider the value function for any state $i \in \Bar{G}$ under policy $\pi$
\begin{align*}
    \Vn^{\pi}(i) = \ & Z(s_m) + \alpha^{m} \C_N(s_m, \pi(s_m)) + \alpha^{m+1} \Biggl( \sum_{j \in \mathcal{L}_0} p_{ij} \Vn^{\pi}(j) \Biggr),
\end{align*}
where $s_m \in \mathcal{L}^{i}_m$, $m \le N$, is the first state from which a sensing action is taken starting from $i$ following $\pi$. Let $B_{\pi}$ denote the Bellman operator corresponding to policy $\pi$, then
\begin{align*}
    B_{\pi}^{m+1}\Vn^*(i) = \ & Z(s_m) + \alpha^{m} \C_N(s_m, \pi(s_m)) + \alpha^{m+1} \Biggl( \sum_{j \in \mathcal{L}_0} p_{ij} \Vn^{*}(j) \Biggr).
\end{align*}
Now observe that
\begin{align*}
    & B_{\pi}^{m+1}\Vn^*(i) - \V^{\pi}(i) =  \alpha^{m+1} \Biggl( \sum_{j \in \mathcal{L}_0} p_{ij} \bigl(\Vn^{*}(j) - \V^{\pi}(j)\bigr) \Biggr) \\
    \implies & \, \Vn^*(i) - \V^{\pi}(i) \leq  \alpha^{m+1} \Biggl( \sum_{j \in \mathcal{L}_0} p_{ij} \bigl(\Vn^{*}(j) - \V^{\pi}(j)\bigr) \Biggr) \\
    & \text{(since \( \Vn^* \leq B_{\pi}^{m+1}\Vn^* \) elementwise)}.
\end{align*}
Note that the above inequality implies~\eqref{eq:Mkn_suboptimality_1}.

To prove~\eqref{eq:Mkn_suboptimality_2}, first note that for any state $j \in \S_n$ we have 
$$
\Vn^*(j) \le \Vn^{\pi_{AS}}(j) = \V^{\pi_{AS}}(j) \le \V^*(j) + \frac{k}{1-\alpha}.
$$
Following similar steps as in the proof of~\eqref{eq:Mkn_suboptimality_1}, for any root state $j \in G,$
\begin{align*}
    & B_{\pi}^{N}\Vn^*(j) - \V^{\pi}(j) \le  \alpha^N(\Vn^*(s_N) - \V^{\pi}(s_N)) \le \frac{\alpha^N k}{1-\alpha},\\
    \implies
    & \, \Vn^*(j) - \V^{\pi}(j) \leq  \frac{\alpha^N k}{1-\alpha},
\end{align*}
where $s_N \in \mathcal{L}^{j}_N$, is the state reached after playing $N$ blind steps starting from $j$ following $\pi$. This establishes~\eqref{eq:Mkn_suboptimality_2}.
\end{proof}
\subsection{Proof of Lemma~\ref{thm:one-step}} 
\label{proof:onestep}
Define $\pi_{N+1}$ as an extension of the policy $\pi^*_{\mathcal{M}_{k,N}}$ for $\mathcal{M}_{k,N+1}$. Without loss of generality (W.L.O.G.), assume that $\pi^*_{\mathcal{M}_{k,N}}(\st) \in A_s$ for all states $\st \in \mathcal{L}_N$. Under $\pi_{N+1}$, states $\st \in \bigcup_{l=0,1,\ldots, N} \mathcal{L}_l$ (i.e., all states from layers $0$ to $N$) are mapped to actions provided by the policy $\pi^*_{\mathcal{M}_{k,N}}$, while states $\st \in \mathcal{L}_{N+1}$ (i.e., states in the $N+1$ layer) are assigned arbitrary actions.

Let $S^j_{exp}$ denote the sequence of states $s \in \mathcal{L}^j_m$ for $m \geq 0$ that are visited under $\pi^*_{\mathcal{M}_{k,N}}$ starting from the root state $j$ (inclusive of $j$). Define $S_{exp} \coloneqq \cup_j S^j_{exp}$. Furthermore, define

$$
Z_{s_m}(s_T) = \sum_{i=m}^{T-1} \alpha^{i-m} \mathcal{B}(s_i) \mathcal{C}(a_i)
$$

for $0 \leq m \leq T-1$, where the only difference from $Z(s_T)$ is that we start from state $s_m$ at $t=0$ and calculate the cumulative cost to reach $s_T$.

\begin{align}
Z(i) + \alpha^{N+1} \min_{a} \Bigl(\mathcal{B}(i)\mathcal{C}(a) + k + \alpha \mathcal{B}(i)\mathcal{T}(a)V^*_{\mathcal{M}_{k,N}} \Bigr) \geq V^*_{\mathcal{M}_{k,N}}(j) \quad \forall j \in \mathcal{S}, \ i \in \mathcal{L}^j_{N+1}.
\label{next}
\end{align}
We claim that \eqref{next} is a necessary and sufficient condition for the optimal actions to remain unchanged for all states \( \st \in S_{exp} \) in every step of policy iteration. This follows from the fact that there exists an improvable action at some state \( \st \in S_{exp} \) in \( \mathcal{L}_m \) for the improved policy \( \pi'_{N+1} \) at some step of the policy iteration algorithm if and only if \eqref{Thm1} is satisfied for some \( i \in \mathcal{L}_{N+1} \) and \( a \in A \).

\begin{align}
    Z_{\st}(i) + \alpha^{N+1-m} \left(\mathcal{B}(i)\mathcal{C}(a) + k + \alpha \mathcal{B}(i)\mathcal{T}(a)V^*_{\mathcal{M}_{k,N}} \right) 
    < V^{\pi'_{N+1}}_{\mathcal{M}_{k,N}}(\st) 
    \label{Thm1}
\end{align}
By inequality \eqref{next}, we have
\begin{align*}
    Z(\st) + Z_{\st}(i) + \alpha^{N+1-m}(\mathcal{B}(i)\mathcal{C}(a) + k + \alpha \mathcal{B}(i)\mathcal{T}(a)V^*_{\mathcal{M}_{k,N}} ) 
\ge Z(\st) + V^*_{\mathcal{M}_{k,N}}(\st).
\end{align*}
Simplifying, we get
\begin{align*}
Z_{\st}(i) + \alpha^{N+1-m}(\mathcal{B}(i)\mathcal{C}(a) + k + \alpha \mathcal{B}(i)\mathcal{T}(a)V^*_{\mathcal{M}_{k,N}} )
\ge V^*_{\mathcal{M}_{k,N}}(\st),
\end{align*}
which is the necessary and sufficient condition for the optimal value function of all root states to remain unchanged even when evaluated on \( \mathcal{M}_{k,N+1} \).
\subsection{Proof of Theorem~\ref{thm:pi_N_optimal}}
\label{proof:pi_N_optimal}
\begin{proof}
    Suppose that~\eqref{ultimate} holds.
    Fix a state \( i \in \mathcal{L}^j_{N+1} \) and consider a policy \( \pi^{ji} \) such that we traverse \( i \) starting from root state \( j \) by following this policy. Let \( \mathcal{M}_0 \) denote the corresponding MDP with no sensing cost. Then,
\begin{align*}
V_{\mathcal{M}_0}^{\pi^{ji}}(j) &\ge Z(i) + \alpha^{N+1} V^*_{\mathcal{M}_0}(i), \\
V_{\mathcal{M}_0}^{\pi^{ji}}(j) &\ge Z(i) + \alpha^{N+1} \min_{a}\left(\mathcal{B}(i)\mathcal{C}(a) + \alpha \mathcal{B}(i)\mathcal{T}(a) V^* \right), \\
V_{\mathcal{M}_0}^{\pi^{ji}}(j) &\ge Z(i) + \alpha^{N+1} V_{AS;0}(i).
\end{align*}
Let \( \pi^j_{M} \) be a policy such that, starting from root state \( j \) and following \( \pi^j_{M} \), we take \( M > N \) consecutive blind steps. Note that
\begin{align}
V_{\mathcal{M}_0}^{\pi^j_{M}}(j) &\ge \min_{i \in \mathcal{L}^j_{N+1}} \left(Z(i) + \alpha^{N+1} V_{AS;0}(i) \right). \label{use}
\end{align}
W.L.O.G., assume \( \pi^*_{\mathcal{M}_{k,N}}(\st) \in A_s \) for all \( \st \in \mathcal{L}_N \). If condition~\eqref{ultimate} holds, then
\begin{align*}
V^*_{\M_{k,N}}(j) &\le \min_{i \in \mathcal{L}^j_{N+1}} \left(Z(i) + \alpha^{N+1} V_{AS;0}(i) \right)
\le V_{\M_0}^{\pi_j^M}(j)
\le V_{\M_k}^{\pi_j^M}(j)\\
\implies V^*_{\M_{k}}(j) &\le V^{\pi^*_{\M_{k,N}}}_{\M_{k}}(j) = V^*_{\M_{k,N}}(j) \le V_{\M_k}^{\pi_j^M}(j)
\end{align*}
Thus, the optimal policy for $\mathcal{M}_k$ takes at most $N$ consecutive blind steps starting from $j$, and consequently, when~\eqref{ultimate} holds, $\Vn(j) = \V(j)$ for all $j \in \mathcal{S}$.

Now consider a scenario where condition~\eqref{ultimate} does not hold and notice that this condition is not satisfied if and only if $\epsilon_N > 0$. Exactly as in the proof of Theorem~\ref{thm:depththreshold} (see Section~\ref{proof:depth}), for a stationary policy $\pi$ of $\mathcal{M}_k$, define a set of root states $G$ such that, starting from any state $j \in G$ and following $\pi$, at least $N+1$ consecutive blind steps are taken. Similarly, define $\Bar{G} := \mathcal{S} \setminus G$.
We have already proved that for root states $j \in G$,
\begin{align}
&V_{\mathcal{M}_k}^{\pi}(j) \ge \min_{i \in \mathcal{L}^j_{N+1}} \left(Z(i) + \alpha^{N+1} V_{AS;0}(i) \right)\notag \\ \implies &V_{\mathcal{M}_{k,N}}^*(j) - V_{\mathcal{M}_k}^{\pi}(j) \le \epsilon_N, \quad \forall i \in G.
\label{eq:Mkn_suboptimality_epsion}
\end{align}
Now, the value function for any state $l \in \Bar{G}$ under policy $\pi$ can be represented as 
\begin{align*}
    \Vn^{\pi}(l) = \ & Z(s_m) + \alpha^{m} \C_N(s_m, \pi(s_m)) + \alpha^{m+1} \Biggl( \sum_{j \in \mathcal{L}_0} p_{lj} \Vn^{\pi}(j) \Biggr),
\end{align*}
where $s_m \in \mathcal{L}^{l}_m$, $m \le N$, is the first state from which a sensing action is taken starting from $l$ following $\pi$. Let $B_{\pi}$ is the Bellman operator corresponding to policy $\pi$, then
\begin{align*}
    B_{\pi}^{m+1}\Vn^*(l) = \ & Z(s_m) + \alpha^{m} \C_N(s_m, \pi(s_m)) + \alpha^{m+1} \Biggl( \sum_{j \in \mathcal{L}_0} p_{lj} \Vn^{*}(j) \Biggr).
\end{align*}
Now observe that
\begin{align*}
    & B_{\pi}^{m+1}\Vn^*(l) - \V^{\pi}(l) =  \alpha^{m+1} \Biggl( \sum_{j \in \mathcal{L}_0} p_{lj} \bigl(\Vn^{*}(j) - \V^{\pi}(j)\bigr) \Biggr) \\
    \implies & \, \Vn^*(l) - \V^{\pi}(l) \leq  \alpha^{m+1} \Biggl( \sum_{j \in \mathcal{L}_0} p_{lj} \bigl(\Vn^{*}(j) - \V^{\pi}(j)\bigr) \Biggr) \\
    & \text{(since \( \Vn^* \leq B_{\pi}^{m+1}\Vn^* \) elementwise)}
\end{align*}
where $p_{lj}$ denotes the probability of landing in root state $j \in \mathcal{S}$ after taking the first sensing step when starting from $l$ and following $\pi$. Consider $G' = \{s \mid s \in \Bar{G} \text{ and } s \not\rightarrow i, \forall i \in G\}$, i.e., starting from any state $s \in G' $  we never reach any state $i \in G$, under policy $\pi$ on $\M_{k,N}$. Therefore, we have
\begin{align*}
  \Vn^*(i)-\V^{\pi}(i) &\le \epsilon_N , \quad \forall i \in G,\\
  \Vn^{*}(i)-\V^{\pi}(i) &= 0 , \quad \forall i \in G',  \\
 \Vn^{*}(i)-\V^{\pi}(i) &\le \alpha^{a_i} \Bigl( \sum_{j \in \mathcal{L}_0} p_{ij}  \bigl(\Vn^{*}(j)-\V^{\pi}(j)\bigr) \Bigr), \quad \forall i \in \Bar{G},
\end{align*}
where $a_i$'s are policy $\pi$ and root state-dependent constants, with $a_i \geq 1$. Identical to the argument made in Section~\ref{proof:depth}, treat \( f(i) := \Vn^*(i) - \V^{\pi}(i) \) as the reward in state \( i \) corresponding to a Markov chain over \( \mathcal{S} \) with transition probabilities \( \{p_{ij}\} \), where the states in \( G \cup G' \) are absorbing. Since the reward on absorbing states is at most~$\epsilon_N$ (see~\eqref{eq:Mkn_suboptimality_epsion}), it follows that $f(i)=\Vn^*(i)-\V^{\pi}(i) \leq \alpha \epsilon_N$ for all $i \in \Bar{G}\setminus G'.$ (for all non-absorbing states). This establishes ~\eqref{eqn:suboptimalitybound}
% The set of equations above represents the dynamic programming equations for a Markov Process with a state space $\mathcal{S}$ and zero transition costs, featuring state-dependent discounting factors. Here, $\Vn^{*}(i)-\V^{\pi}(i)$ denotes the value function for the state $i$, and $G \cup G'$ denotes the set of absorbing states. Also, note that any non-absorbing state reaches an absorbing state a.s. (hitting time is finite). Since $a_i \ge 1$ for all $i \in G$, we can conclude that $|\Vn^{*}(i)-\V^{\pi}(i)| \leq \alpha \epsilon_N$ for all $i \in \Bar{G}\setminus G'$ (for all non-absorbing states).

\textbf{NOTE:}\\
\textbf{1.} Even if condition~\eqref{eq:more-ultimate} is satisfied and the optimal policy for a root state \( j \) of \( \mathcal{M}_k \) is restricted to having a maximum of \( N \) consecutive blind steps, it does not generally imply that \( V_{\mathcal{M}_k}^*(j) = V_{\mathcal{M}_{k,N}}^*(j) \).

\textbf{2.} It follows from the same proof that the stronger claim below holds for any root state \( j \):
\begin{align*}
    \hspace{-8pt} V^*_{\mathcal{M}_k}(j)
    \ge \min \Bigl\{ \min_{i \in \mathcal{L}^j_{N+1}} \hspace{-6pt} \left(Z(i) + \alpha^{N+1} V_{AS;0}(i) \right), V^*_{\mathcal{M}_{k,N}}(j) - \alpha \hspace{-2pt} \max_{s \in \mathcal{S} \setminus \{j\}} \left[ V^*_{\mathcal{M}_{k,N}}(s) - \hspace{-6pt}\min_{r \in \mathcal{L}^s_{N+1}} \hspace{-5pt} \left(Z(r) + \alpha^{N+1} V_{AS;0}(r) \right) \right]^+ \Bigr\}.  
\end{align*}
Here, $[x]^+$ denotes the positive part of $x.$

\noindent
\textbf{Idea:} Define a separate terminal value for each of the states \( j \in G \), given by
\begin{align*}
\Vn^*(j) - V_{\mathcal{M}_k}^{\pi}(j) \le \Vn^*(j) - \min_{i \in \mathcal{L}^j_{N+1}} \left(Z(i) + \alpha^{N+1} V_{AS;0}(i) \right).
\end{align*}
\end{proof}
\subsection{Proof of Lemma~\ref{thm:epsilon}}
\label{proof:epsilon}
\begin{proof}
Let the baseline MDP $\mathcal{M}$ be defined according to the conditions specified in the lemma. For any state $\tilde{s} \in \mathcal{S}_{\infty}$, we have
\begin{align}
    V_{AS;0}(\tilde{s}) &= \mathcal{B}(\tilde{s}) \mathcal{C}(\pi_{AS}(\tilde{s})) + \alpha \mathcal{B}(\tilde{s}) \mathcal{T}(\pi_{AS}(\tilde{s})) V^*,
    \label{eqn:def_Vas}
\end{align}
where $\mathcal{B}(\tilde{s}) = \mathcal{B}(\tilde{s}) \mathcal{T}(\pi_{AS}(\tilde{s}))$.
We claim that for the above non-trivial MDP, for each $a \in A$, there exists a root state $r$ such that
\begin{align}
    V^*(r) &< \mathcal{C}(r,a) + \alpha e_r \mathcal{T}(a) V^*.
    \label{eq:root_inequality}
\end{align}
Also note that $\exists \ N^*$ s.t. $\forall N \ge |\mathcal{S}| - 1$, every element of $\mathcal{B}(\tilde{s})$ is non-zero $\forall \tilde{s} \in \mathcal{L}_N$. Hence, by applying the inequality from~\eqref{eq:root_inequality}, we obtain
\begin{align}
\mathcal{B}(\st) V^* &< \mathcal{B}(\tilde{s}) \mathcal{C}(a) + \alpha \mathcal{B}(\st) \mathcal{T}(a) V^* \quad \forall a \in A, \notag\\
\implies \mathcal{B}(\tilde{s}) V^* &< V_{AS;0}(\tilde{s}).
\label{eqn:ineq_Vas}
\end{align}
Thus, for all $\st \in \mathcal{L}_{N}$, where $N \ge |\mathcal{S}| -2 $, applying~\eqref{eqn:ineq_Vas} to the definition of $V_{AS;0}(\st)$ in~\eqref{eqn:def_Vas}, we obtain
\begin{align}
V_{AS;0}(\st) < \mathcal{B}(\st)\mathcal{C}(a)+ \alpha V_{AS;0}(\st_a) \quad \forall a \in A
\label{eqn:intermediate}
\end{align}
Now let $i'$ be the state reached after taking a blind step with action $a$ from state $i \in \mathcal{L}_{N}$. Then, it immediately follows from~\eqref{eqn:intermediate} that
\begin{align*}
Z(i) + \alpha^{N} V_{AS;0}(i) <  Z(i) + \alpha^{N} \left(\mathcal{B}(i)\mathcal{C}(a) + \alpha V_{AS;0}(i')\right)
% Z(i) + \alpha^{N}  \left(\mathcal{B}(i)\mathcal{C}(a) + \alpha V_{AS;0}(i')\right)\\
= Z(i') + \alpha^{N+1} V_{AS;0}(i').
% < Z(i') + \alpha^{N+1} (\mathcal{B}(i') \mathcal{C}(b)+ \alpha   V_{AS;0}(i'_b)) \quad \forall b \in A
\end{align*}
% \begin{multline*}
% Z(i) + \alpha^{N}  \left(\mathcal{B}(i)\mathcal{C}(a) + \alpha V_{AS;0}(i')\right)\\
% = Z(i') + \alpha^{N+1} V_{AS;0}(i') \\
% < Z(i') + \alpha^{N+1} (\mathcal{B}(i') \mathcal{C}(b)+ \alpha   V_{AS;0}(i'_b)) \quad \forall b \in A
% \end{multline*}
% where $\mathcal{B}(i'_b) = \mathcal{B}(i') \mathcal{T}(b)$.
Thus
\begin{align*}
\min_{i \in \mathcal{L}^{j}_{N}} \left(Z(i) + \alpha^{N} V_{AS;0}(i) \right) &< \min_{i \in \mathcal{L}^{j}_{N+1}} \left(Z(i) + \alpha^{N+1} V_{AS;0}(i) \right)\\
\epsilon_{N+1} &< \epsilon_{N}.
\end{align*}
To prove $\epsilon_{N+1} \le \epsilon_{N},$  for a general MDP, we simply replace the strict inequalities in~\eqref{eqn:ineq_Vas} and~\eqref{eqn:intermediate} with non-strict ones.
\end{proof}

%  \item[(ii)]
% Note that for such an MDP for any policy $\pi$ and state $i \in \mathcal{L}_N$, we have
%     $$
%     % V_{\M_0}^{\pi}(i) \ge \mathcal{B}(i)V_{\M_k}^* >V_{blind}( \mathcal{B}(i))
%     $$
%     Hence Inequality \eqref{use} is replaced with a strict inequality, i.e.,
% \begin{align*}
% V_{\M_0}^{\pi^j_{M}}(j)  &> \min_{i \in \mathcal{L}^{j}_{N+1}} E_{N+1}(j,i)\\
%  \delta_N& < \epsilon_N
% \end{align*} 

% \end{enumerate}
\section{Experimental Design and Setup}
\label{sec:Experimental Design}
{\bf Experimental Setup:} Experiments were conducted on a MacBook Air with an Apple M3 chip and 16GB of memory. Hyperparameters were set according to the defaults in \texttt{POMDPs.jl} \cite{juliapomdps}, with adjustments clearly stated and made to ensure comparable performance or runtime. For SARSOP, the reward and sensing values were scaled by a factor of 1000 to match performance on the Frozen Lake task, and the policy computation time was increased from 1s to 100s for the Taxi task. It is important to note that this scaling was applied only during the policy computation phase using SARSOP with its default hyperparameters. The resulting policy was then evaluated in the original, unscaled environment, consistent with the evaluation of SPI and other methods, which used unscaled values throughout. Without this scaling, SARSOP performed poorly on the task—even relative to $\pi_{\text{ATM}}$ and FIB. 

{\bf Initialization of POMDP Baselines:} The POMDP planning algorithms have been initialized with a belief distribution corresponding to the starting state distribution, rather than a uniform distribution over all states. For example, in Frozen Lake, where the starting state is deterministic, the initialized belief is a degenerate distribution concentrated entirely on the initial state. In contrast, for Stochastic Taxi, the initialized belief is uniform over the 300 valid states from which the initial state is uniformly sampled.

{\bf Choice of Domain:} 
Our choice of the ICU-Sepsis benchmark environment \cite{icu} is motivated by \citet{nam2021reinforcement}, who employ a similar tabular sepsis simulator to model personalized treatment strategies for ICU patients under state observation costs.
ICU-Sepsis is constructed from real medical data (MIMIC-III database) and serves as a standardized benchmark for evaluating RL algorithms.  

We also incorporate gridworld-based domains inspired by \citet{bellinger2021active}, who utilize standard RL Gym environments such as FrozenLake 8$\times$8 (see Table~\ref{table:8x8}) and Taxi to benchmark RL algorithms for MDPs with state-sensing costs.
While \citet{nam2021reinforcement} \& \citet{bellinger2021active} focused on learning-based RL algorithms, we adapted these domains to evaluate planning algorithms.

To evaluate performance in smaller state spaces, we conducted experiments on both the standard Frozen Lake 4×4 environment and a custom-designed “hard” variant, as shown in Tables~\ref{table:4x4 default} and~\ref{table:4x4 hard}, respectively. The results from the custom hard setup revealed that $\pi_{\text{ATM}}$ and FIB may struggle to find effective policies even in small state spaces when faced with challenging planning problems, thus highlighting the robustness of SPI in such scenarios.

We also conducted experimental analysis on other non-grid world-based domains, such as an Inventory Management. This case study, detailed in the Appendix~\ref{Inventory}, features two distinct sensing cost scenarios. We benchmarked our algorithm against $\pi_{\text{ATM}}$, the truncated MDP approach, and utilized ~\ref{adv:bound} to derive suboptimality bounds on their performance.

We highlight that we tested nearly all offline \texttt{POMDPs.jl} planners (Incremental Pruning, PBVI, QMDP, etc) but excluded them from the experimental setup due to prohibitive runtimes/ineffective policies.

\textbf{Setting hyperparameters for SPI.} The SPI heuristic relies on two interpretable hyperparameters: $maxsteps$, which limits the length of blind action sequences explored by PolicyUpdate for each root state $s$, and $\delta$, the improvement threshold that governs the termination of PolicyUpdate iterations. Below, we outline practical guidelines for setting these parameters.

   $\bullet$ The choice of $maxsteps$ is dependent on the desired precision of the Value function. We recommend choosing $maxsteps$ such that $\alpha^{maxsteps} \cdot k$ is smaller than the desired precision (or one order of magnitude smaller). For instance, in the ICU-Sepsis environment  ($\alpha = 0.99$, maximum sensing cost from Table 1 being $k = 0.1$), since we report and compare values only up to 3 decimal places, we set $maxsteps = 500$ as $0.99^{500} \cdot 0.1 \approx 6.6 \times 10^{-4} < 10^{-3}$.
   
   $\bullet$ The improvement threshold $\delta$ controls the number of iterations in PolicyUpdate, terminating when the value function improvement falls below $\delta$ for all root states. We suggest setting $\delta$ to be one order of magnitude below the desired precision. Alternatively, the PolicyUpdate iterations in SPI can be capped at a fixed number (e.g., 4–5), as SPI converges rapidly. For instance, in the ICU-Sepsis task, the improvement falls below $10^{-4}$ after two iterations, and for other benchmark environments, it falls below $10^{-6}$ in 3–4 iterations.
\subsection{Performance of POMCP on Frozen Lake}
\label{POUCT}
In the case of Frozen Lake, the belief distribution for all the POMDP solvers was solely concentrated on the single starting state \textbf{S}. Despite this initialization, the PO-UCT algorithm performed significantly worse than SPI. For instance, in the Frozen Lake 4x4 (Hard) environment, even when the starting state was positioned in the bottom-right corner, closer to the goal (see Table~\ref{table:4x4 modified hard}), the PO-UCT algorithm—with over 290k rollouts ($\approx$ 500 sec) per planning step—achieved an average return of $1.91 \times 10^{-2}$. In contrast, SPI and $\pi_{\text{ATM}}$ achieved average returns of $5.44 \times 10^{-2}$ and $4.7 \times 10^{-2}$, respectively. Moreover, as the starting state moved farther from the goal, the expected return of PO-UCT declined sharply compared to SPI, likely due to PO-UCT's poor performance with sparse rewards, which requires significantly more planning time.

\begin{table}[t]
\centering
\begin{tabular}{cccc}
  \begin{subtable}[t]{0.23\textwidth}
    \centering
    \scalebox{0.9}{
      \begin{tabular}{c@{}c@{}c@{}c}
        \textcolor{yellow}{S} & F & F & F \\
        F & \textcolor{red}{H} & F & \textcolor{red}{H} \\
        F & F & F & \textcolor{red}{H} \\
        \textcolor{red}{H} & F & F & \textcolor{green}{G} \\
      \end{tabular}
    }
    \caption{Default 4×4 grid}
    \label{table:4x4 default}
  \end{subtable}
  &
  \begin{subtable}[t]{0.23\textwidth}
    \centering
    \scalebox{0.9}{
      \begin{tabular}{c@{}c@{}c@{}c}
        F & \textcolor{red}{H} & \textcolor{yellow}{S} & F \\
        F & \textcolor{green}{G} & \textcolor{red}{H} & F \\
        F & \textcolor{red}{H} & \textcolor{red}{H} & F \\
        F & F & F & F \\
      \end{tabular}
    }
    \caption{Custom-hard 4×4 grid}
    \label{table:4x4 hard}
  \end{subtable}
  &
  \begin{subtable}[t]{0.23\textwidth}
    \centering
    \scalebox{0.9}{
      \begin{tabular}{c@{}c@{}c@{}c@{}c@{}c@{}c@{}c}
        \textcolor{yellow}{S} & F & F & F & F & F & F & F \\
        F & F & F & F & F & F & F & F \\
        F & F & F & \textcolor{red}{H} & F & F & F & F \\
        F & F & F & F & F & \textcolor{red}{H} & F & F \\
        F & F & F & \textcolor{red}{H} & F & F & F & F \\
        F & \textcolor{red}{H} & \textcolor{red}{H} & F & F & F & \textcolor{red}{H} & F \\
        F & \textcolor{red}{H} & F & F & \textcolor{red}{H} & F & \textcolor{red}{H} & F \\
        F & F & F & \textcolor{red}{H} & F & F & F & \textcolor{green}{G}
      \end{tabular}
    }
    \caption{Default 8×8 grid}
    \label{table:8x8}
  \end{subtable}
  &
  \begin{subtable}[t]{0.23\textwidth}
    \centering
    \scalebox{0.9}{
      \begin{tabular}{c@{}c@{}c@{}c}
        F & \textcolor{red}{H} & F & F \\
        F & \textcolor{green}{G} & \textcolor{red}{H} & F \\
        F & \textcolor{red}{H} & \textcolor{red}{H} & F \\
        F & F & F & \textcolor{yellow}{S} \\
      \end{tabular}
    }
    \caption{Custom-hard 4×4 with start state \textbf{S} closer to goal}
    \label{table:4x4 modified hard}
  \end{subtable}
\end{tabular}
\caption{\footnotesize \textbf{Frozen Lake grid environments} used in our experiments.}
\label{table:FrozenLakeInstance}
\end{table}

\section{Analysis of SPI}
\label{sec:SPI Analysis}

\subsection{Performance Guarantees of SPI}
$\mathsf{PolicyUpdate}$ guarantees that the output policy $\pi_o$ is at least as "good" as the input reference policy $\pi_{ref}$ for all root states ($\pi_o \overset{\mathcal{S}}{\succeq} \pi_{ref}$).
Consequently, the updated policy $\pi_{improv}$, resulting from each application of $\mathsf{PolicyUpdate}$ within SPI, is at least as effective as the previous policy for all root states and reduces the value function by $\delta$ for at least one root state at each iteration until termination. We can, therefore, derive a theoretical upper bound on the number of $\mathsf{PolicyUpdate}$ steps.

SPI is guaranteed to terminate for any $\delta>0$ due to the assumption of a bounded cost function and a finite state space. Specifically, since $\mathsf{PolicyUpdate}$ produces a policy $\pi_{\text{improv}}$ such that $\pi_{\text{improv}} \overset{\mathcal{S}}{\succeq} \pi'$, and the value function cannot decrease by more than $\delta$ for some state indefinitely—due to the optimal value function being lower bounded—the algorithm must eventually converge.

Moreover, if the initial policy $\pi_{\text{init}}$ is chosen to be the Always-Sense (AS) policy—as in our numerical experiments (Section~\ref{sec:numerics})—then SPI is guaranteed to terminate within at most $\frac{k |\mathcal{S}|}{\delta (1 - \alpha)}$ iterations. This follows from the fact that the AS policy is suboptimal by at most $\frac{k}{1 - \alpha}$ for any state, and each $\mathsf{PolicyUpdate}$ step decreases the value function by at least $\delta$ for some state.
In practice, however, SPI typically converges in far fewer iterations—even in large state spaces. For instance, in the ICU-Sepsis task, the improvement in return fell below $10^{-4}$ after just two iterations.
\subsection{Adapting SPI to Continuous State Spaces}
\label{sec:adapt}
SPI can be naturally extended to continuous state spaces using function approximation, an avenue we find promising for future research. Below, we provide an informal sketch of how $\mathsf{PolicyUpdate}$ (and consequently SPI) can be adapted to continuous state spaces (with finite action space):
\begin{enumerate}
    \item 
Our policy could be represented as a neural network with a final softmax activation layer, mapping the state features of the last sensed root state, concatenated with the sequence of blind actions played thereafter, to a probability distribution over the action space. Thus, the belief state features are defined as the root state features of the corresponding root state concatenated with the sequence of blind actions.
\item
We first note that the $\mathsf{PolicyUpdate}$ algorithm, in all its steps, requires the value function to be evaluated for the reference policy only at the root states. Therefore, it is sufficient to have a critic (value function) neural network that is accurate at the root state features. Since we are in a planning setup, we have access to the transition probability function and reward functions, allowing us to train the critic to fit the value function corresponding to $\pi_{ref}$ at the root state features by rolling out the reference policy $\pi_{ref}$ at root states.
\item
Since we are dealing with a continuous state space setup, we replace Line 2 of the $\mathsf{PolicyUpdate}$ algorithm to sample states from some distribution over the root states (e.g., root state visit distribution of $\pi_{ref}$), because it is not feasible to carry the procedure for all root states in a continuous state space.
\item
An estimate of the value of $\mathcal{B}(\tilde{s})\mathcal{C}(a)$ at a belief state $\tilde{s}$ for action $a$, which is necessary in Lines 9 and 10 and for evaluating $V_{MS}$, can be easily obtained using the transition function and cost function.
\item
In this setting, Lines 12 and 15 correspond to fitting the candidate policy $\pi'$ by using the belief state features as input to produce the corresponding suitable action as output. Similarly, Line 21 updates the output policy $\pi_o$ in the same manner.
For the comparison between the value functions performed in Line 19, a critic network for $\pi'$ can be trained using an approach similar to that described above for $\pi_{ref}$.
\end{enumerate}
The only change required in the SPI algorithm is in Line 3, where we could simply terminate after a fixed number of iterations.
\section{Analysis of ATM Heuristic}
\label{sec:ATM Analysis}
\begin{algorithm}[tb]
\caption{Act-Then-Measure Heuristic (ATM)}
\label{alg:policy_improvement_heuristic}
\textbf{Input}: $\tilde{s} \in \mathcal{S}_\infty$ (state) \\
\textbf{Output}: Action to play $\tilde{a} \in \mathcal{A}_\infty$
\begin{algorithmic}[1]
\State $a = \pi_{AS}(\tilde{s})$
\If{$\alpha \Big( V_{AS}\big(\mathcal{B}(\tilde{s})\mathcal{T}(a)\big) - \mathcal{B}(\tilde{s})\mathcal{T}(a)V^* \Big) < \dfrac{k}{1-\alpha}$}\label{algo:heuristic-check}
    \State \textbf{Play} $(a, blind)$
\Else
    \State \textbf{Play} $(a, sense)$
\EndIf
\end{algorithmic}
\end{algorithm}
In this section, we provide an alternative perspective for understanding and analyzing the ATM heuristic~\citep{ATM}, adapted to our expanded state-space framework. The proposed heuristic, formalized in Algorithm~\ref{alg:policy_improvement_heuristic}, operates as follows. Given a state $\tilde{s}$, it compares the (expected discounted) cost of following the AS policy with that of taking the one-time blind action $(\pi_{AS}(\tilde{s}), \bl)$ and subsequently following the AS policy. If the former cost is less, the algorithm selects the sensing action $(\pi_{AS}(\tilde{s}), \s)$, i.e., it follows the action prescribed by AS; otherwise, it executes the blind action $(\pi_{AS}(\tilde{s}), \bl)$.

We demonstrate that the ATM heuristic, in fact, arises from a policy improvement step; consequently, the following guarantee holds.
\begin{theorem}
    The policy described by Algorithm~\ref{alg:policy_improvement_heuristic} dominates any always-sensing policy.
\end{theorem}
\begin{proof}
It suffices to show that Line~\ref{algo:heuristic-check} in Algorithm~\ref{alg:policy_improvement_heuristic} constitutes a policy improvement step. For any~$a \in A,$
\begin{align*}
Q_{AS}(\st,(a,\s)) &= B(\st) \C(a) + \alpha B(\st) \T(a) V^* + \frac{k}{1-\alpha}, \\
Q_{AS}(\st,(a,\bl)) &= B(\st) \C(a) + \alpha V_{AS}(B(\st)\T(a)).
\end{align*}
It is therefore easy to check that 
\begin{align*}
& Q_{AS}(\st,(a,\bl)) < Q_{AS}(\st,(a,\s))   \\
\iff & \alpha \big( V_{AS}(\mathcal{B}(\st)\mathcal{T}(a)) - \mathcal{B}(\st)\mathcal{T}(a)V^* \big) < \frac{k}{1-\alpha}.
\end{align*}
\end{proof}

\section{Additional Results}
\label{sec:addl-results}
\subsection{Extension of Theorem~\ref{thm:pi_N_optimal}}
\label{sec:extension}
Theorem~\ref{thm:pi_N_optimal} establishes a sufficient condition~\eqref{ultimate} for the optimal policy of $\M_{k,n}$ to also be optimal for $\M_k$. However, this condition~\eqref{ultimate} must hold for all root states for the result to apply.

A stronger result follows from the proof of Theorem~\ref{thm:pi_N_optimal}: If, for any root state~$j,$ it holds that 
\begin{equation}
\label{eq:more-ultimate}
  Z(i) + \alpha^{N+1} V_{AS;0}(i) \ge V^*_{\M_{k,N}}(j) \quad \forall\ i \in \mathcal{L}^j_{N+1},\end{equation}
then the optimal policy for $\M_k$ takes at most $N$ consecutive blind steps starting from~$j.$ Thus, if \eqref{eq:more-ultimate} is satisfied at certain root states, one does not need to explore depths~$N+1$ and beyond at these root states.

Finally, as the following lemma shows, the suboptimality bound $\epsilon_N$ is decreasing in~$N.$
\begin{lemma}
\label{thm:epsilon}
It always holds that $\epsilon_{N+1} \le \epsilon_{N},$ where $\epsilon_N$ is as defined in the statement of Theorem~\ref{thm:pi_N_optimal}. Furthermore, if $\M$ is irreducible, and there exists no action that is optimal for all states in the baseline MDP, then $\epsilon_{N+1} < \epsilon_{N}$ for all  $N\ge |S|-2.$
\end{lemma}

Additionally, Theorem~\ref{thm:pi_N_optimal} provides a computable upper bound on the suboptimality of the policy $\pi^*_{\M_{k,N}}$ for $\M_k$. Building on its proof in Appendix~\ref {proof:pi_N_optimal}, we derive a stronger result for computing a lower bound on $V^*_{\M_k}$: For any root state $j$,
\begin{align}
    \hspace{-8pt} V^*_{\mathcal{M}_k}(j)
    \ge \min \Bigl\{ \min_{i \in \mathcal{L}^j_{N+1}} \hspace{-6pt} \left(Z(i) + \alpha^{N+1} V_{AS;0}(i) \right), V^*_{\mathcal{M}_{k,N}}(j) - \alpha \hspace{-2pt} \max_{s \in \mathcal{S} \setminus \{j\}} \left[ V^*_{\mathcal{M}_{k,N}}(s) - \hspace{-6pt}\min_{r \in \mathcal{L}^s_{N+1}} \hspace{-5pt} \left(Z(r) + \alpha^{N+1} V_{AS;0}(r) \right) \right]^+ \Bigr\}.  
    \label{adv:bound}
\end{align}
Here, $[x]^+$ denotes the positive part of $x.$

\textbf{Note:} The lower bound on $V^*_{\M_k}$ in Figure~\ref{fig:II} was computed using the stronger bound provided above.

\subsection{Counter-example related to Lemma~\ref{thm:one-step}} 
\label{sec:counterexample}
We begin with an example that demonstrates that starting at a root state, if a certain policy~$\pi$ is optimal for $\M_{k,N}$ as well as $\M_{k,N+1},$ that does not guaranteee that the $\pi$ is also optimal for~$\M_k.$ Consider the two-state two-action baseline MDP shown in Figure~\ref{fig:countermdp}, with sensing cost~$k = 0.005$ and discount factor~$\alpha = 1/2.$

\begin{figure}[t]
\centering
\begin{tikzpicture}[state/.style={draw, circle, minimum size=0.75cm},>=Stealth,node distance=3cm]
\definecolor{neublue}{RGB}{33,113,181}   % professional deep blue
\definecolor{neured}{RGB}{200,33,33}     % professional deep red
% Nodes
\node[state] (0) {0};
\node[state, right=of 0] (1) {1};

% Edges for Action Red
\draw[->, neured, thick, dashed] (0) to[bend right=20] node[midway, below]{$0.72$, $0.066$} (1); % Red
\draw[->, neublue, thick] (0) to[bend right=70] node[midway, below]{$0.69$, $0.29$} (1);         % Blue

\draw[->, neured, thick, dashed] (1) to[bend left=-20] node[midway, above]{$0.934$, $0.502$} (0); % Red
\draw[->, neublue, thick] (1) to[bend left=-70] node[midway, above]{$0.481$, $0.41$} (0);         % Blue

% Loops
\draw[->, neured, thick, dashed] (0) to [out=-135,in=135,loop,looseness=4.8] node[below=1.3cm of 0] {$0.28$, $0.066$} (0); % Red
\draw[->, neublue, thick] (0) to [out=-120,in=120,loop,looseness=10.8] node[above=1.3cm of 0] {$0.31$, $0.29$} (0);        % Blue

\draw[->, neured, thick, dashed] (1) to [out=45,in=-45,loop,looseness=4.8] node[below=1.3cm of 1] {$0.066$, $0.502$} (1); % Red
\draw[->, neublue, thick] (1) to [out=60,in=-60,loop,looseness=10.8] node[above=1.3cm of 1] {$0.519$, $0.41$} (1);        % Blue

\end{tikzpicture}
\caption{\footnotesize \textbf{Two-state two-action MDP} with actions $\{Red, Blue\}$, $k =0.005$ and $\alpha = 0.5$}
\label{fig:countermdp}
\end{figure}
The optimal policy and value function corresponding to the MDPs~$\M_{k,N}$ for different choices of $N$ are tabulated in Table~\ref{tab-counter1}. The optimal policy is shown as the sequence of actions to take starting at any root state, terminating in a sensing action; for example, `BRRRR' means to take the sequence of blind actions `BRRR' and then the sensing action `R.' Note that for $N \leq 3,$ the optimal policy at root states for $\M_{k,N}$ is to take a sensing action. However, for $N \geq 4,$ is optimal to make a sequence of blind steps in root state~$1$. 
\ignore{This is further illustrated in Figure~\ref{fig:countermdp-2}, where we see that taking a blind action in root state~1 becomes increasingly attractive for increasing $N,$ until it becomes optimal at $N=4.$} %killed to save space

Interestingly, we find in this example that the criterion for Theorem~\ref{thm:pi_N_optimal} (see ~\ref{sec:extension}) is satisfied for root state $0$ at $N=2$. Therefore, it is clear at that point that the optimal policy for root state $0$ will take a maximal of $2$ blind steps, and we can restrict our search for the optimal policy starting at $0$ until the $2^{\text{nd}}$ layer. The same criterion is \emph{not} satisfied for root state $1$ for $N \le 6.$

\begin{table}[t]
    \centering
    \small % Use a smaller font size
    \renewcommand{\arraystretch}{1.1}
    \begin{tabular}{|c|p{1.4cm}|p{1.4cm}|p{1.4cm}|p{1.4cm}|}
    \hline
    $N$ & $\pi^*_{\M_{k,N}}(0)$ & $V^*_{\M_{k,N}}(0)$ & $\pi^*_{\M_{k,N}}(1)$ & $V^*_{\M_{k,N}}(1)$ \\ \hline
    0 & R & $0.367061$ & B & $0.6796465$ \\ \hline
    1 & R & $0.367061$ & B & $0.6796465$ \\ \hline
    2 & R & $0.367061$ & B & $0.6796465$ \\ \hline 
    3 & R & $0.367061$ & B & $0.6796465$ \\ \hline 
    4 & R & $0.36703456$ & BRRRR & $0.67958256$ \\ \hline 
    5 & R & $0.367029$ & BRRRRR & $0.6795691$ \\ \hline 
    6 & R & $0.3670226$ & BRRRRRR & $0.6795541$ \\ \hline
    \end{tabular}
    \caption{\footnotesize \textbf{Change in optimal policy and value function} with the number of blind steps for baseline MDP in Figure~\ref{fig:countermdp}}
    \label{tab-counter1}
\end{table}

\section{Extension to Non-Uniform Sensing Costs}
\label{sec:non-uniform}
Sensing costs can often be context-dependent and may depend on resource constraints or belief uncertainty. The constraint of non-uniform sensing costs can be easily incorporated in our setup (for both the truncated MDP approach as well as the SPI heuristic along with the theoretical results) by modelling the sensing cost as state-action-dependent for the augmented MDP, i.e., $k': \mathcal{S}_{\infty} \times A \rightarrow \mathbb{R}$
. For example, in the case where the sensing cost depends on belief uncertainty, the sensing cost can simply be treated as a function of the state in the augmented MDP since a state (of the augmented MDP) directly relates to the belief distribution over the root states. More formally, with such constraints,
\begin{enumerate}
    \item \textbf{Modification to Truncated MDP approach:} The cost function $\C_{\infty}$ associated with $\M_{k'}$ will be suitably modified as follows
\begin{align*}
    \C_{\infty}(\tilde{s}, (a,\s)) &= \B(\tilde{s})\C(a) + k'(\tilde{s}, a)\\
    \C_{\infty}(\tilde{s}, (a,\bl)) &= \B(\tilde{s})\C(a)
\end{align*}
This will, in turn, lead to the modification of the cost function $\C_n: \mathcal{S}_n \times \A_n \rightarrow \mathbb{R}$ associated with the truncated MDP $\M_{k',n}$. The rest of the truncated MDP parameters remain unchanged, and we can directly compute its optimal policy and value function for the truncated MDP.
\item  \textbf{Modification to SPI:} The myopic sensing value function $V_{MS}$
 and its corresponding policy $\pi_{MS}$ will also depend on the state and will be modified as follows:
\begin{align*}
V_{MS}(\B(\st), \Bar{V}) = \min_{a \in A} \bigl(B(\st) \C(a) + \alpha B(\st) \T(a) \Bar{V} + k'(s,a) \bigr)\\
\pi_{MS}(\B(\st), \Bar{V}) = \arg \min_{a \in A} \bigl(B(\st) \C(a) + \alpha B(\st) \T(a) \Bar{V} + k'(s,a)\bigr)
\end{align*}
With this simple modification, SPI extends to scenarios with non-uniform sensing costs.
\item \textbf{Modification to Theoretical Results:} As with the truncated MDP  approach and SPI, our results can be similarly extended to scenarios with non-uniform sensing costs, with simple modifications to its proof. For instance, Theorem~\ref{thm:depththreshold} is modified as follows:
\[
V_{\mathcal{M}_{k',N}}^* (j) - V_{\mathcal{M}_{k'}}^*(j) \le  \frac{\alpha^N \tilde{k}}{1-\alpha} \quad \forall j \in \mathcal{S}, N\ge 0,
\]
where 
\begin{align*}
\tilde{k} &= \sup_{a \in A, \tilde{s} \in \mathcal{S}_{\infty} \setminus \mathcal{L}_{N-1}} k'(\tilde{s}, a) \quad \text{for } N \ge 1, \\
\tilde{k} &= \sup_{a \in A, \tilde{s} \in \mathcal{S}_{\infty}} k'(\tilde{s}, a) \quad \text{for } N = 0.
\end{align*}

\end{enumerate}

\section{Inventory Management Case Study}
\label{Inventory}
% {\bf An inventory management example:} 
This example is adapted from \cite{bradley1977applied}. We consider an inventory with a capacity of $3$ units. The demand for items is either $1$ or $2$ units, each with probability~1/2 at every step (month). The production cost for an item is $\$1000$ per unit, while the selling price stands at $\$2000$ per unit, ensuring a profit of $\$1000$ units per sale. We consider a holding cost of $\$500$ on each month for each remaining item in the inventory by month-end.\footnote{Holding cost is evaluated based on the no. of remaining items of the inventory at the end of the month after meeting the demand.} Furthermore, consider a sensing cost of either $\$200$ or $\$64$ for observing the remaining items in the inventory (the state), and we aim to maximize the discounted profit with the discounting factor $\alpha=0.8$.

\begin{figure}[t]
\centering
\begin{subfigure}[b]{0.48\textwidth}
    \centering
    \includegraphics[width = \linewidth]{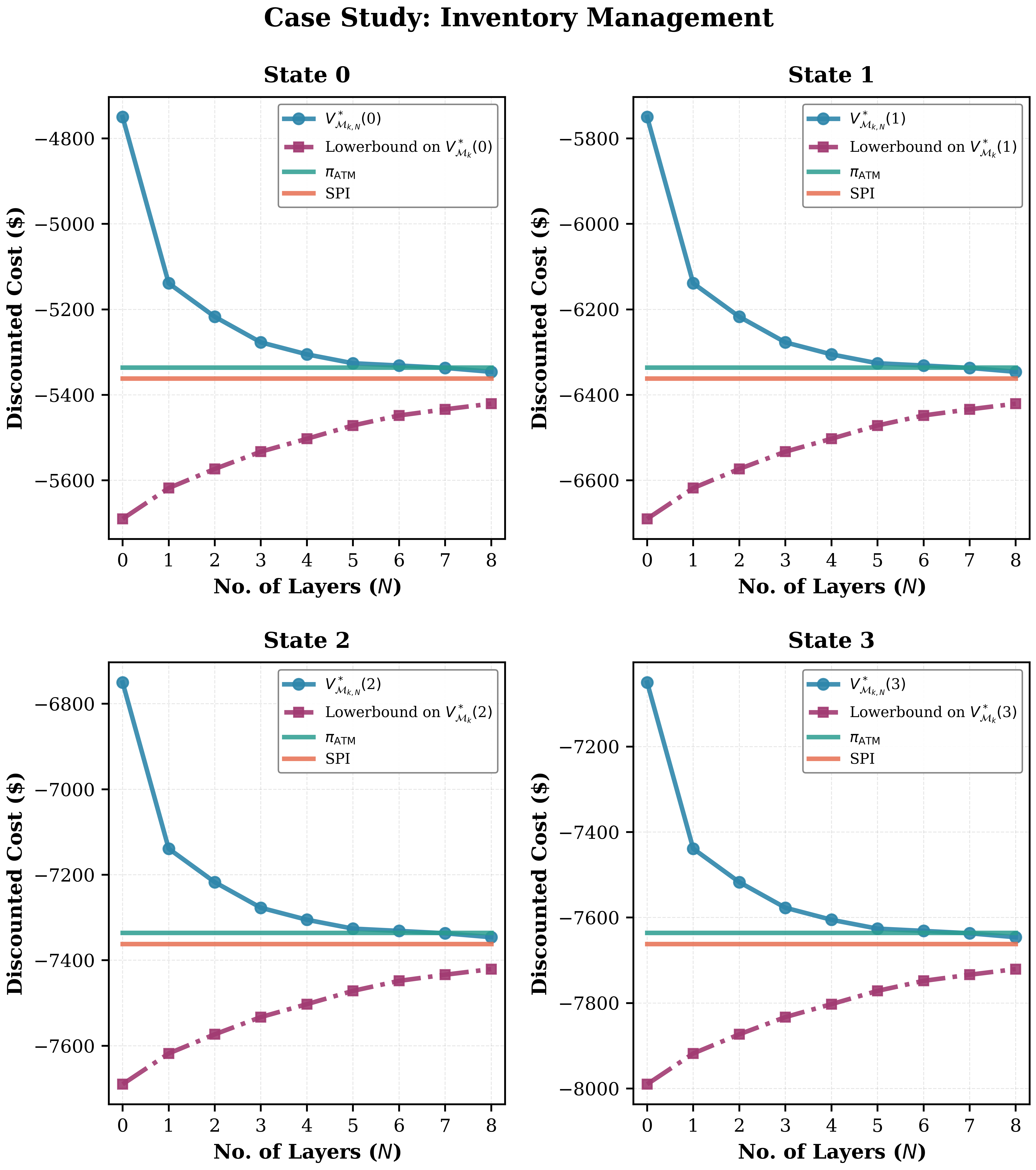}
    \caption{Inventory management with sensing cost \$200}
    \label{fig:inventory_200}
\end{subfigure}
\hfill
\begin{subfigure}[b]{0.48\textwidth}
    \centering
    \includegraphics[width = \linewidth]{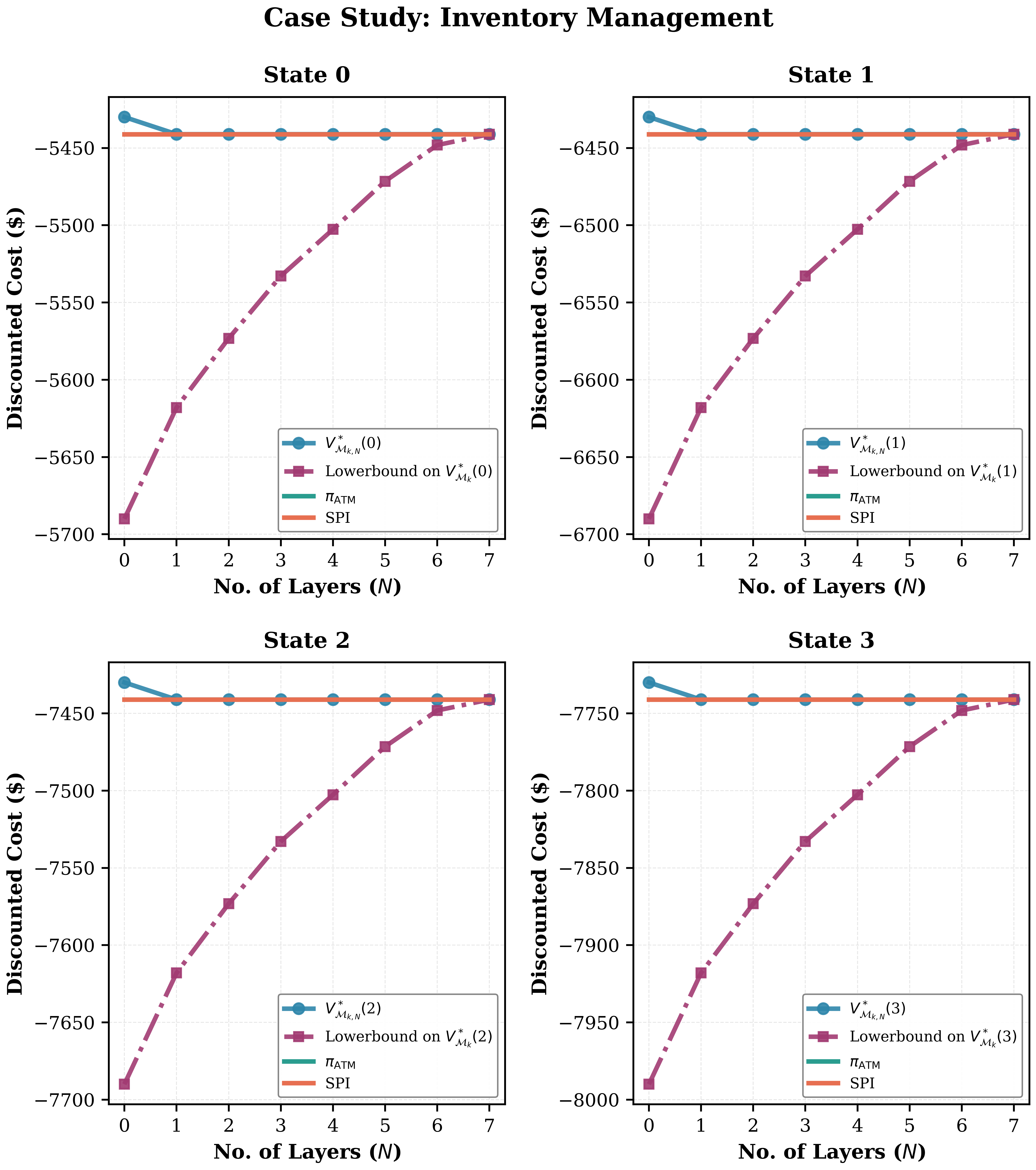}
    \caption{Inventory management with sensing cost \$64}
    \label{fig:inventory_64}
\end{subfigure}
\caption{\footnotesize \textbf{Applying Thm~\ref{thm:pi_N_optimal}} to Inventory Management}
\label{fig:inventory_comparison}
\end{figure}

Our main takeaways are as follows. For sensing cost~\$200, our results are shown in Figure~\ref{fig:inventory_200}. In this case, we see that SPI (Section~\ref{sec:heuristic}) performs quite close to the optimal policy (judging by the bound on sub-optimality gap) and the optimal policy for the truncated MDP $\M_{k,N}$ outperforms it only after $N \ge 7$. However, the conditions of Theorem~\ref{thm:pi_N_optimal} are not satisfied over the depths $N$ we were able to compute for (recall that the computational complexity of solving~$\M_{k,N}$ grows exponentially in~$N$). This is consistent with the results in Figure~\ref{fig:inventory_200}; we continue to see small cost benefits from increasing the threshold on the number of blind actions allowed.

For the lower sensing cost of \$64, our results are shown in Figure~\ref{fig:inventory_64}. In this case, we see that the heuristic policy (Section~\ref{sec:heuristic}), which does provide an improvement over always sensing, is in fact optimal for~$\M_k$. Moreover, the optimal policy for $\M_{k,1}$ is also found to be optimal for~$\M_k$. However, the condition of Theorem~\ref{thm:pi_N_optimal} is only satisfied at $N=7.$ 

\end{document}